\documentclass[10pt,twocolumn,letterpaper]{article}

\usepackage{cvpr}              %
\usepackage[accsupp]{axessibility} %

\usepackage{times}
\usepackage{epsfig}
\usepackage{graphicx}
\usepackage{amsmath}
\usepackage{amssymb}
\usepackage{enumitem}
\usepackage{multirow}
\usepackage{xspace}
\usepackage{booktabs}  %
\usepackage{arydshln}  %
\usepackage[table,dvipsnames]{xcolor}  %
\usepackage{soul}  %
\usepackage[most]{tcolorbox}
\usepackage{pifont}
\usepackage{nicematrix}  %
\usepackage{makecell}
\usepackage{subcaption}

\usepackage{kotex}

\usepackage[dvipsnames]{xcolor}
\newcommand{\red}[1]{{\color{red}#1}}

\newcommand{\green}[1]{{\color{ForestGreen}#1}}
\newcommand{\orange}[1]{\textcolor{Peach}{#1}}
\newcommand{\blue}[1]{\textcolor{RoyalBlue}{#1}}

\newcommand{\avgn}{Avg$^\text{N}$\xspace}
\newcommand{\mmbdev}{MMB\xspace}
\newcommand{\mmep}{MME$^{\text{P}}$\xspace}
\newcommand{\seedimg}{SEED$^{\text{I}}$\xspace}
\newcommand{\llavaw}{LLaVA$^{\text{W}}$\xspace}
\newcommand{\model}{Honeybee\xspace}

\newcommand{\cabs}{C-Abstractor\xspace}
\newcommand{\dabs}{D-Abstractor\xspace}

\newcommand{\cmark}{\ding{51}}
\newcommand{\xmark}{\ding{55}}

\newcommand{\modelXinst}{Honeybee$_\text{\ w/o Inst.}$\xspace}
\newcommand{\modelOinst}{Honeybee$_\text{\ w/ Inst.}$\xspace}

\newcommand{\tmpl}[1]{\{\textbf{#1}\}}
\newcommand{\tmplR}[1]{\{\textbf{\textcolor{red}{#1}}\}}

\definecolor{mygreen}{HTML}{3cb44b}

\definecolor{cvprblue}{rgb}{0.21,0.49,0.74}

\newcommand{\mat}[1]{\mathbf{#1}} %

\newcolumntype{x}[1]{>{\centering\arraybackslash}p{#1pt}}
\newcommand{\tablestyle}[2]{\setlength{\tabcolsep}{#1}\renewcommand{\arraystretch}{#2}\centering\small}

\definecolor{light-gray}{gray}{0.50}
\newcommand{\gray}[1]{\textcolor{gray}{#1}}

\usepackage[pagebackref,breaklinks,colorlinks,citecolor=cvprblue]{hyperref}

\def\paperID{8009} %
\def\confName{CVPR}
\def\confYear{2024}

\title{Honeybee: Locality-enhanced Projector for Multimodal LLM}

\author{Junbum Cha\thanks{Equal contribution}
\and
Wooyoung Kang\footnotemark[1]
\and
Jonghwan Mun\footnotemark[1]
\and
Byungseok Roh
\vspace{0.01cm}
\and
Kakao Brain\\
{\tt\small \{junbum.cha, edwin.kang, jason.mun, peter.roh\}@kakaobrain.com}
}

\begin{document}
\maketitle

\begin{abstract}
In Multimodal Large Language Models (MLLMs), a visual projector plays a crucial role in bridging pre-trained vision encoders with LLMs, enabling profound visual understanding while harnessing the LLMs' robust capabilities. Despite the importance of the visual projector, it has been relatively less explored. In this study, we first identify two essential projector properties: ($i$) flexibility in managing the number of visual tokens, crucial for MLLMs' overall efficiency, and ($ii$) preservation of local context from visual features, vital for spatial understanding. Based on these findings, we propose a novel projector design that is both flexible and locality-enhanced, effectively satisfying the two desirable properties. Additionally, we present comprehensive strategies to effectively utilize multiple and multifaceted instruction datasets. Through extensive experiments, we examine the impact of individual design choices. Finally, our proposed MLLM, Honeybee, remarkably outperforms previous state-of-the-art methods across various benchmarks, including MME, MMBench, SEED-Bench, and LLaVA-Bench, achieving significantly higher efficiency. Code and models are available at \url{https://github.com/kakaobrain/honeybee}.
\end{abstract}

\section{Introduction}
\label{sec:intro}

Large Language Models (LLMs) have made great progress in recent years, mainly thanks to instruction tuning. Visual instruction tuning~\cite{llava} has been proposed to extend LLMs into Multimodal LLMs (MLLMs) to perceive and understand visual signals (\eg, images). The main idea for MLLMs is to introduce a projector connecting the vision encoder and LLM, and to learn the projector using visual instruction data while keeping the parameters of the vision encoder and LLM. Such a simple technique allows to preserve and leverage the pre-trained knowledge and abilities in vision encoder and LLM, making resulting MLLMs unlock new capabilities, such as generating stories, poems, advertisements, code, and more from given images; those tasks have traditionally been considered challenging for conventional vision-language foundation models~\cite{coca,florence}. Such success leads to increasing attention for research into MLLMs taking multimodal inputs (\eg, videos~\cite{videochat}, audio~\cite{pengi}, 3d world~\cite{3dllm}, point cloud~\cite{pointllm}) beyond text.

For MLLMs, the projector plays a critical role in the following two aspects: 
1) \textit{performance}: as it bridges the vision and language models by translating visual features into visual tokens so that the language model can understand, the quality of conveyed visual tokens directly impacts the overall performance of the MLLM;
and 2) \textit{efficiency}: as most of the computational burden lies with the language model, the efficiency of MLLMs is heavily influenced by the number of resulting visual tokens.
However, despite its critical importance, the projector has been relatively underexplored in the literature and most MLLMs simply adopt either linear projectors~\cite{llava,shikra} or abstractors~\cite{mplug,Qwen-VL,instructBLIP,minigpt,blip2}.

\begin{figure}[!t]
    \begin{center}
    \scalebox{0.98}{
        \includegraphics[width=\linewidth]{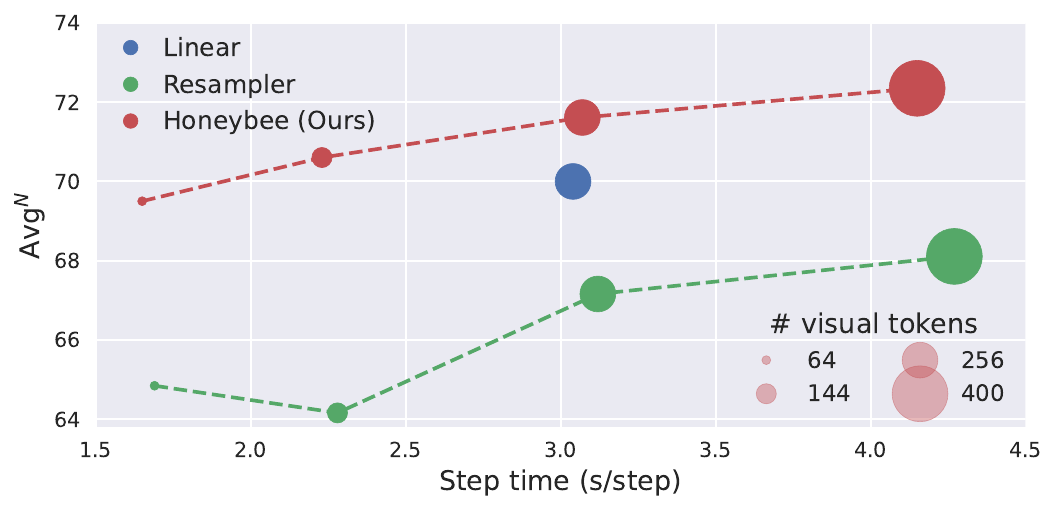}
    }
    \vspace{-1.1cm}
    \end{center}
    \caption{
         \textbf{Performance \textit{vs.} efficiency for projectors} where \avgn means an average of normalized benchmark scores (MME, MMBench, and SEED-Bench) and step time is a single step execution time during pre-training.
         \model with the locality-enhanced projector (\ie, C-Abstractor) offers a more favorable balance between efficiency and performance over existing projectors.
    }
    \label{fig:tradeoff}
    \vspace{-0.2cm}
\end{figure}

\begin{table}[!t]
    \tablestyle{2pt}{1.05}
    \begin{center}
    \scalebox{0.86}{
        \begin{tabular}{@{}l|ccccc@{}}
        \toprule
         & \mmbdev & \seedimg & \mmep & MME & \llavaw \\ \midrule
        Previous SoTA   & {67.7 \cite{llava-v1.5}} & {68.1 \cite{llava-v1.5}} & {1531 \cite{llava-v1.5}} & {1848 \cite{Qwen-VL}} & {70.7 \cite{llava-v1.5}} \\
        \textbf{\model (Ours)} & \textbf{73.6} \scriptsize{(\green{+5.9})} & \textbf{68.6} \scriptsize{(\green{+0.5})} & \textbf{1661} \scriptsize{(\green{+130})} & \textbf{1977} \scriptsize{(\green{+129})} & \textbf{77.5} \scriptsize{(\green{+6.8})} \\ 
        \bottomrule
        \end{tabular}
    }
    \end{center}
    \vspace{-0.55cm}
    \caption{\small
        \textbf{Comparison with SoTA.} The proposed \model outperforms the previous state-of-the-art MLLMs on various benchmarks with significant gaps.
    }
    \vspace{-0.3cm}
    \label{table:teaser-comp-w-sota}
\end{table}

Notably, recent MLLMs prefer abstractors (\eg, resampler, Q-former) to linear projectors; this is primarily due to their flexibility in handling the number of resulting visual tokens, thus offering versatile design options for achieving a preferable balance between efficiency and performance. However, as shown in \cref{fig:spatial-understanding}, the abstractors face more difficulties in learning spatial understanding tasks compared to the linear projectors. This difficulty stems from the abstraction process lacking a locality-aware design, which causes it to primarily focus on a few regions, leading to a loss of finer details essential for spatial comprehension. In contrast, linear projectors excel at preserving all local contexts of visual features via one-to-one transformation. This strong preservation of locality allows effective spatial understanding.

Motivated by this, we propose novel locality-enhanced projectors, which exhibit a more favorable balance between performance (by locality preservation) and efficiency (by abstraction capability) as presented in \cref{fig:tradeoff}. To be specific, we introduce two locality-enhanced projectors by employing two powerful operations in locality modeling---convolution and deformable attention. Such injection of locality-aware design into the abstraction process not only promotes the overall performance improvement of MLLMs in handling intricate visual information but also capitalizes on computational efficiency during the subsequent response generation phase of LLMs.

On top of the MLLM with a locality-enhanced projector, named \textit{\model}, we offer a hidden recipe for cutting-edge MLLMs. Notably, a prevalent strategy in recent MLLM training involves multiple instruction data: 1) GPT-assisted instruction-following dataset like LLaVA~\cite{llava} and 2) vision-language task datasets with \textit{instructization}\footnote{Instructization denotes conversion of raw data into instruction-following format using pre-defined templates.} process~\cite{instructBLIP}. To take maximized advantage from these datasets, we present important but less explored design choices for 1) how to utilize multifaceted instruction data and 2) the effective way for an instructization process. We perform extensive experiments to verify the impact of individual design choices on diverse benchmarks and hope to offer valuable insights into training strong MLLMs.

Our main contributions are summarized as follows:
\begin{itemize}[leftmargin=.3cm,noitemsep,nosep]
    \item 
        We identify two crucial properties of projector, 1) locality preservation of visual features and 2) flexibility to manage the number of visual tokens, and propose locality-enhanced abstractors to achieve the best of both worlds.
    \item
        We propose a (hidden) effective way to tackle multifaceted datasets as well as the instructization process, maximizing the benefit from instruction data.
    \item 
        With the locality-enhanced projector and explored hidden recipes, our \model achieves state-of-the-art performances across the various MLLM benchmarks---MME, MMBench, SEED-Bench, and LLaVA-Bench (\Cref{table:teaser-comp-w-sota}).
\end{itemize}

\section{Related Work}
\label{sec:related_work}

\subsection{Multimodal Large Language Models}
\label{rel:mllm}

The remarkable instruction-following and generalization abilities of recent LLMs have ushered in extending LLMs to Multimodal LLMs (MLLMs).
Early works such as Flamingo~\cite{flamingo} and BLIP-2~\cite{blip2} successfully adapted LLMs to visual tasks, showing notable zero-shot generalization and in-context learning capabilities.
More recently, MLLMs are further advanced mainly through visual instruction tuning, which includes utilizing vision-language (VL) datasets~\cite{Qwen-VL,lynx,instructBLIP} and enhancing visual instruction-following data~\cite{llava,minigpt,svit,llavar,empirical,lrv}.
Also, several studies focus on grounding capabilities of MLLMs by utilizing additional datasets specifically designed for these tasks~\cite{ferret,shikra,kosmos,pink}.
However, recent MLLMs have not yet deeply explored visual projectors, despite the proper design of projectors is critical in both the effectiveness and efficiency of MLLMs.

\subsection{Multimodal Instruction-following Data}
\label{rel:data}
The breakthrough from GPT-3~\cite{gpt3} to ChatGPT~\cite{chatgpt} highlights the importance of instruction-following data in empowering LLM to understand and follow natural language instructions.
Similarly, integrating visual instruction data is essential for training MLLMs to handle various instructions, thus increasing their versatility.
Several studies employ a powerful LLM, \eg, GPT-4~\cite{gpt4}, to generate visual instruction data for complex VL tasks, such as generating stories, poems, detailed captions from given images~\cite{llava,llavar,minigpt,svit,lrv}. 
Another line of studies has explored transforming existing VL task datasets into an instruction-following format using pre-defined templates, called \textit{instructization} \cite{instructBLIP,lynx,llava-v1.5,Qwen-VL}. 
While there is active development and expansion of instruction-following datasets, the research focusing on how to combine and utilize these datasets remains underexplored.

\begin{figure*}[!ht]
    \begin{center}
    \scalebox{0.85}{
        \includegraphics[width=\linewidth]{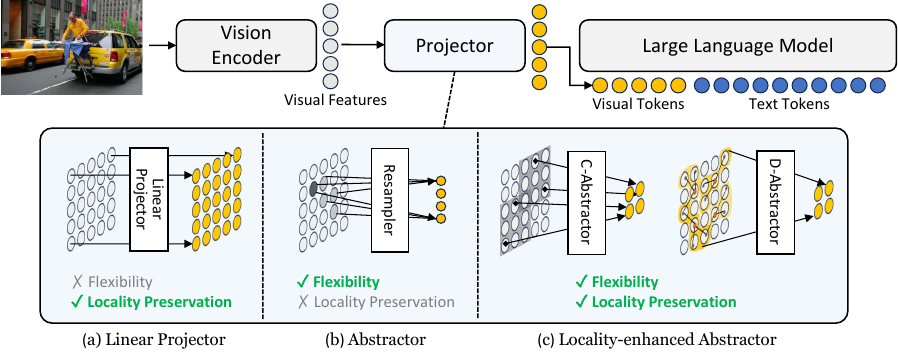}
    }
    \vspace{-0.7cm}
    \end{center}
    \caption{
         \textbf{Conceptual comparison between projectors} in terms of how to convert visual features into visual tokens.
         (a) Linear projector performs a one-to-one transformation, thus effective in preserving all local contexts of visual features, but limited in flexibility.
         (b) Abstractor such as resampler offers flexibility by abstracting the visual features into a smaller number of visual tokens but is limited in local context preservation by focusing on salient regions.
         (c) Our locality-enhanced abstractors can achieve both flexibility and locality preservation.
    }
    \vspace{-0.3cm}
    \label{fig:comp_projectors}
\end{figure*}

\subsection{Benchmarks for MLLM}
\label{rel:benchmark}
MME~\cite{mme}, MMBench~\cite{mmb}, and SEED-Bench~\cite{seed} have been introduced as comprehensive benchmarks for the \textit{objective evaluation} of MLLMs with yes/no or multiple-choice questions. These benchmarks encompass a broad spectrum of evaluation tasks, ranging from coarse- and fine-grained perceptual analysis to visual reasoning tasks.
On the other hand, as the capabilities of MLLMs evolve to handle more complex VL tasks such as visual storytelling and instruction-following in an open-set manner with free-form text, other types of benchmarks have been proposed, \ie, \textit{subjective evaluation}. Following NLP studies~\cite{llmeval1,llmeval2}, several studies leverage powerful LLMs, \eg, GPT-4~\cite{gpt4}, to assess the response quality of MLLMs~\cite{llava,touchstone,mmvet}. This approach aims for a more detailed evaluation of the proficiency of MLLMs.
In this paper, we aim to provide valuable insights into training a robust and high-performing MLLM through extensive analysis.

\section{\model: Locality-enhanced MLLM}
\label{sec:method}

\subsection{Overview}
Generally, the goal of Multimodal Large Language Models (MLLMs) is to learn a model that can produce instruction-following responses for the given multimodal inputs.
In this paper, we consider images as an additional modality input to MLLMs.
Thus, the language model becomes a receiver of both visual and text (instruction) tokens while generating text responses in an autoregressive manner. Formally, a multimodal input consists of two types of tokens: image tokens $\mat{X}_{\texttt{img}}$ and text tokens $\mat{X}_{\texttt{text}}$.
Then, the language model predicts the response $\mat{Y}=\{w_i\}_{i=1}^{L}$ conditioned on the multimodal input where $L$ means the number of tokens in the response.
Therefore, the response is predicted by
\begin{equation}
    p(\mat{Y}|\mat{X}_{\texttt{img}}, \mat{X}_{\texttt{text}}) = \prod_{i=1}^{L}p(w_i|\mat{X}_{\texttt{img}}, \mat{X}_{\texttt{text}}, w_{<i}).
    \label{eq:mllm_loss}
\end{equation}

\vspace{-0.65cm}
\paragraph{Architecture.}
MLLMs are generally composed of three networks: 1) \textit{vision encoder}, 2) \textit{projector}, and 3) \textit{large language model (LLM)}.
The vision encoder provides a sequence of region-level visual features for detailed image understanding.
The projector is in charge of transferring the visual features to visual tokens for the subsequent language model.
Then, the LLM processes the fused visual and instruction tokens and produces a response autoregressively.

\vspace{-0.35cm}
\paragraph{Efficiency of MLLMs.}
In the MLLM architecture, the LLM predominantly accounts for the entire computation and memory consumption of the MLLM. Thus, with the same LLM, the efficiency of the MLLM---in terms of computation, memory consumption, and throughput---is mainly affected not by the efficiency of the visual encoder and projector, but by the number of resulting visual tokens fed into the LLM. This is also shown in \cref{fig:tradeoff} and \cref{sec:appendix:efficiency-of-mllms}.

\vspace{-0.35cm}
\paragraph{Revisiting existing projectors.}
The projector takes the $N$ visual features and converts them into $M$ visual tokens.
For the projector, MLLMs adopt an operation between a linear projection and an abstraction of visual features.
The linear projection is simple yet effective, particularly in preserving knowledge and understanding of vision encoder (\eg, the locality of visual features), but faces challenges in scalability and efficiency, primarily due to its inherent constraint of one-to-one transformation between visual features and tokens (\ie, $M = N$).
On the other hand, the abstraction offers a more adaptable approach to determining the quantity of visual tokens ($M$).
For example, resampler and Q-former utilize $M$ (generally $<N$ for efficiency) learnable queries and cross-attention to extract visual cues from visual features \cite{mplug,Qwen-VL,instructBLIP,minigpt,flamingo}.
While such flexibility by abstraction allows better efficiency, but it can inherently suffer from a risk of information loss from the vision encoder.

\subsection{Locality-enhanced Projector}
In this section, we first describe our motivation for locality-enhanced projectors.
Then, we present two types of locality-enhanced projectors (\cabs and \dabs) and describe the training pipeline.

\vspace{-0.3cm}
\subsubsection{Motivation}
\label{subsubsec:motivation}
\vspace{-0.2cm}
The projector is crucial as it bridges visual and language models, translating image features into a format that is comprehensible and utilizable by the language model. Considering its role, when designing a projector, the most important factor is flexibility in deciding the number of resulting visual tokens. As described above, the number of visual tokens produced by the projector determines the overall efficiency and computational amount of MLLM. Considering the scenario of handling multiple or large images, improving efficiency through flexibility in reducing the number of visual tokens is highly required for scalability.
This requirement has led to the preference for abstractors like resamplers and Q-formers over linear projectors in recent MLLMs~\cite{Qwen-VL,instructBLIP,blip2,mplug}.

\begin{figure}[!t]
    \begin{center}
    \scalebox{0.95}{
        \includegraphics[width=\linewidth]{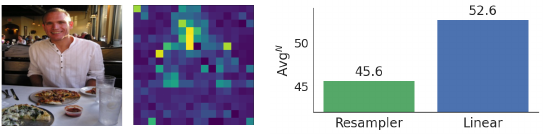}
    }
    \vspace{-0.6cm}
    \end{center}
    \caption{
         (Left) an example of an attention map from the resampler and (Right) a comparison of spatial understanding capability for the resampler and linear projector where \avgn is computed using six spatial understanding tasks from MME, \mmbdev, and \seedimg.
    }
    \vspace{-0.1cm}
    \label{fig:spatial-understanding}
\end{figure}

However, we observe the resampler suffers from tackling spatial understanding tasks compared to the linear projector. Note that a linear projector retains all the local context of visual features through a one-to-one projection without loss. In contrast, in \cref{fig:spatial-understanding}, the resampler tends to summarize information primarily from a few regions (\eg, man) while potentially overlooking details in some local regions (\eg, meals, cups, background people). We believe that this difference between two models in the preservation of all local contexts (during abstraction) significantly impacted spatial understanding performance.

Stemming from these observations, we propose two novel visual projectors, \cabs and \dabs, under two key design principles: ($i$) enabling flexibility over the number of visual tokens and ($ii$) effectively preserving the local context.
These new projectors are designed to maintain the strengths of the abstractor, such as computational efficiency via flexibility in managing visual token numbers, while also improving the preservation of local features. This enhancement not only boosts the overall performance of MLLMs in handling complex visual information but also benefits from the computational efficiency during the subsequent response generation phase of LLMs.
The conceptual comparison between the existing and proposed projectors is illustrated in \cref{fig:comp_projectors}.

\vspace{-0.2cm}
\subsubsection{Architecture}

\begin{figure}[t]
    \begin{center}
    \scalebox{0.90}{
        \includegraphics[width=\linewidth]{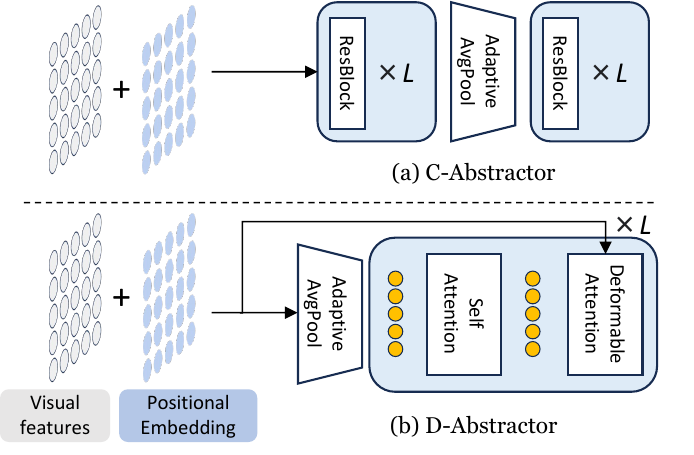}
    }
    \vspace{-0.5cm}
    \end{center}
    \caption{
         Conceptual architecture of our proposed projectors.
    }
    \vspace{-0.1cm}
    \label{fig:arch}
\end{figure}

\paragraph{\cabs.} 
In deep learning, convolution has been the most successful architecture for modeling local context \cite{lecun1995convolutional,resnet-bottleneck,vgg}. Thus, we design \underbar{C}onvolutional \underbar{Abstractor}, \cabs, for effective local context modeling. \cref{fig:arch}a depicts the entire architecture, comprising $L$ ResNet blocks~\cite{resnet-bottleneck} followed by adaptive average pooling and another $L$ ResNet blocks. This design allows to abstract visual features to any squared number of visual tokens, and even project to more visual tokens than the original number of visual features. 
We also tested several variants~\cite{convnext,vgg} in \cref{sec:appendix:details-on-projectors}, but ResNet~\cite{resnet-bottleneck} shows the best performance.

\vspace{-0.35cm}
\paragraph{\dabs.}
While convolution is a successful concept in local context modeling, one can argue that it introduces overly strict inductive biases for locality.
Hence, we propose \underbar{D}eformable attention-based \underbar{Abstractor}, \dabs, enhancing the locality-awareness of the resampler during abstraction while keeping its flexibility.
Specifically, the deformable attention~\cite{ddetr} benefits in preserving local context; each learnable query gathers visual features via a 2-D coordinate-based sampling process using reference points and sampling offsets focusing on near the reference points.
Here, we propose an advanced initialization method of reference points where the reference points are manually initialized, distributing uniformly over the whole feature map. This additional technique allows \dabs to capture fine-grained and comprehensive information for a given image.
More detailed explanations are given in \cref{sec:appendix:details-on-projectors}.

\subsection{Training}

We train \model in the two-stage pipeline. In the first stage, we freeze the vision encoder and LLM, focusing on training the proposed locality-enhanced projector. In the second stage, we train both the projector and LLM to enhance deeper visual understanding and generation abilities.

\vspace{-0.4cm}
\paragraph{Pre-training for vision-language alignment.}
\label{sec:pretrain}
The goal of pre-training is to learn a newly introduced visual projector to build connections between the vision encoder and LLM. Using the image-text data (\eg, BlipCapFilt~\cite{blip}, COYO~\cite{COYO}), the pre-training enables MLLM to develop a nuanced understanding of how visual cues align with textual descriptions. During pre-training, the vision encoder and LLM are frozen to keep the fundamental understanding already established in vision and language models.

\vspace{-0.4cm}
\paragraph{Visual instruction tuning.}
\label{sec:finetune}
After the pre-training of the projector for vision-language alignment, in the second stage, we jointly train the projector and LLM to enhance instruction-following capabilities and achieve a more profound visual understanding. For instruction-following, we utilize two GPT-assisted instruction-following datasets, LLaVA~\cite{llava} and ShareGPT~\cite{vicuna}. In addition, to enhance visual understanding, we instructize a wide range of existing datasets using templates, as listed in \Cref{table:datasets}. 
Specifically, our approach includes: 1) employing a range of tasks such as open-ended VQA \cite{vqav2,gqa,ocrvqa,vsr}, multiple-choice VQA \cite{sqa,aokvqa}, captioning \cite{COYO,blip}, and referring expression comprehension (visual grounding and grounded captioning) \cite{vg,refcoco,refcocog,refcocop}; 2) using multiple datasets for each task; 3) applying a fine-grained but single template for each dataset. Detailed examples and descriptions are in \cref{sec:appendix:template-details}.
We thoroughly explore template-based instructization strategies and the utilization of multifaceted datasets in \Cref{sec:hidden_recipe}.

\begin{table}[!t]
    \tablestyle{2pt}{1.05}
    \begin{center}
    \scalebox{0.75}{
        \begin{tabular}{@{}l|l|c@{}}
        \toprule
        Task & Datasets  & \#samples  \\
        \midrule
        Captioning  & BlipCapFilt~\cite{blip}, COYO100M~\cite{COYO}         & 200M   \\
        VQA (Open)  & VQAv2~\cite{vqav2}, GQA~\cite{gqa}, OCRVQA~\cite{ocrvqa}, VSR~\cite{vsr}       & 2.2M  \\
        VQA (MC)    & ScienceQA~\cite{sqa}, A-OKVQA~\cite{aokvqa}             & 0.03M  \\
        REC         & RefCOCO~\cite{refcoco}, RefCOCO+~\cite{refcocop}, RefCOCOg~\cite{refcocog}, VG~\cite{vg}   & 5.7M   \\
        Instruction & LLaVA150K~\cite{llava}, ShareGPT~\cite{vicuna}          & 0.2M   \\
        \bottomrule
        \end{tabular}
    }
    \end{center}
    \vspace{-0.55cm}
    \caption{
        List of all training datasets. 
    }
    \vspace{-0.5cm}
    \label{table:datasets}
\end{table}

\section{Hidden Recipe for Visual Instruction Tuning}
\label{sec:hidden_recipe}
In \Cref{sec:method}, we examine the limitations of current projectors and propose methods for enhancing locality. However, a clear recipe for training cutting-edge Multimodal LLMs (MLLMs) remains unclear. While it is widely known that instruction tuning using existing datasets with the template-based instructization is beneficial \cite{Qwen-VL,llava-v1.5,instructBLIP}, the details of the instructization process are still underexplored---questions persist regarding dataset selection, utilization, and combination strategies. In this section, we aim to clarify these aspects via following the five research questions: ($i$) To what extent does each dataset contribute to the performance of specific tasks? ($ii$) What is an effective balancing strategy between diverse datasets? ($iii$) What is the appropriate granularity for the templates? ($iv$) How significant is the diversity of the templates? ($v$) Do conversation-like multi-turn templates provide additional benefits? 

\vspace{-0.35cm}
\paragraph{Dataset combination.}
In recent MLLM studies, a diverse range of datasets has been employed for training powerful MLLMs~\cite{lynx,Qwen-VL,instructBLIP,minigptv2,llava-v1.5}. This prevalent practice, however, is not accompanied by comprehensive analysis to identify which datasets are critical for specific tasks.
To offer an in-depth analysis of this, we design a systematic ablation experiment. As outlined in Table~\ref{table:datasets}, we categorize the datasets into several task groups. Then, we examine the variations in benchmark performances by sequentially excluding each task group during instruction tuning. Through these ablation experiments, we hope to offer valuable insights into the key factors for design choice regarding the dataset combination.

\vspace{-0.4cm}
\paragraph{Dataset balancing.}
While a wide range of datasets are available for training MLLMs, their sizes differ substantially, as shown in \Cref{table:datasets}.
Also, when training MLLMs, it is common practice to restrict the number of training iterations to preserve the knowledge of a pre-trained LLM.
Consequently, properly balancing the training datasets is crucial to maximize learning diverse skills within the short training schedule.
To examine this, we compare five different balancing strategies: 1) \textit{per-dataset}: uniform sampling for each dataset, 2) \textit{per-task}: uniform sampling for each task, 3) \textit{per-sample-100k}: uniform sampling for each sample with clipping the maximum size of each dataset to 100k \cite{flan}, 4) \textit{per-dataset-tuned}: empirically tuned balancing based on per-dataset strategy.

\vspace{-0.4cm}
\paragraph{Template granularity.} 
While the use of pre-defined templates for transforming existing datasets into an instruction format is widely recognized \cite{flan,lynx,llava-v1.5,instructBLIP}, the appropriate granularity for applying these templates is not clearly established. We design the experiments to compare two approaches with different template granularity: 1) \textit{fine-grained}: applying unique templates for each dataset \cite{flan}, and 2) \textit{coarse-grained}: applying the shared templates across datasets within the same task category \cite{instructBLIP,llava-v1.5}.

\vspace{-0.4cm}
\paragraph{Template diversity.}
Prior to the emergence of GPT-assisted conversation datasets, securing template diversity was critical, often achieved by employing a range of diverse pre-defined templates alongside input inversion strategies\footnote{Input inversion is a task augmentation strategy by reversing input and target, \eg, inversion of VQA generating questions from image and answer.}~\cite{flan2023,lynx,flippedvqa}. However, the introduction of GPT-assisted datasets has seemingly diminished the emphasis on the diversity of templates \cite{llava-v1.5}. The exact role and significance of employing multiple templates and input inversion techniques in the context of GPT-assisted datasets remain less understood. To investigate this, we compare three distinct approaches utilizing: 1) a single template, 2) multiple templates, and 3) multiple templates with input inversion.

\vspace{-0.4cm}
\paragraph{Multi-turn template.}
When utilizing existing datasets, it's common to find multiple input-target pairs for a single image, as seen in VQA datasets with several QA pairs per image. The multi-turn strategy merges these pairs into a single, conversation-like multi-turn example. 
However, this approach can merge semantically overlapped input-target pairs into one example, potentially encouraging simplistic shortcuts in finding answers, particularly in the autoregressive training of MLLMs. 
To mitigate this, we introduce an additional de-duplication strategy, which removes semantically duplicate input-target pairs from the multi-turn examples, thereby preventing shortcut training. We detail this strategy with examples in \cref{sec:appendix:template-details}.

\section{Experiments}
\label{sec:experiments}

\subsection{Settings}

\vspace{-0.1cm}
\paragraph{Benchmarks.}
We adopt four benchmarks specifically designed for Multimodal LLM (MLLM) evaluation, including MME~\cite{mme}, MMBench~\cite{mmb}, SEED-Bench~\cite{seed} and LLaVA-Bench (In-the-Wild)~\cite{llava}.
The first three assess various capabilities of MLLMs, such as perceptual understanding and visual reasoning, using binary yes/no questions (MME) or multiple-choice questions (MMBench, SEED-Bench).
Note that we use splits of MME with perception tasks (\mmep), MMBench-dev (\mmbdev), and SEED-Bench Image-only (\seedimg), respectively. 
Our focus on perception tasks in MME are explained in \cref{sec:appendix:benchmark-characteristics}.
On the other hand, LLaVA-Bench (In-the-Wild), \llavaw, exploits GPT-4 to assess MLLM's descriptive responses, providing a comprehensive view of the model's performance in natural language generation and human preference.

\vspace{-0.4cm}
\paragraph{Metrics.}
We report the official metrics computed using official implementation for individual benchmarks by default; we also report the normalized average \avgn~\cite{uniter,value} across benchmarks, defined as the average of scores normalized by their respective upper bound scores, facilitating straightforward comparisons.

\vspace{-0.4cm}
\paragraph{Implementation details.}
We use 7B and 13B Vicuna-v1.5~\cite{vicuna} as LLM.
We leverage the pre-trained CLIP ViT-L/14 \cite{CLIP} with 224 and 336 resolutions for 7B- and 13B-LLM, respectively; we use features from the second-last layer of CLIP instead of the last layer. Any image indicator tokens, \eg, special tokens enclosing visual tokens, are not used. We train the entire LLM instead of parameter-efficient fine-tuning.
For in-depth ablations, we use a short training schedule (50k pre-training, 4k instruction tuning) with Vicuna-7B, CLIP ViT-L/14, and \cabs with $M$=144 visual tokens unless stated otherwise. For the final models, we adopt a long training schedule (200k pre-training, 10k instruction tuning). More details are in \cref{sec:appendix:impl-details}.

\vspace{-0.1cm}
\subsection{Analysis on Locality-Enhanced Projector}
\begin{table}[!t]
    \tablestyle{3pt}{1.05}
    \begin{center}
    \scalebox{0.85}{
                
        \begin{tabular}{@{}l|l|cc|cccccc|c@{}}
        & \multirow{2}{*}{Projector} & \multirow{2}{*}{$M$} & \multirow{2}{*}{s/step} & MME & \multicolumn{3}{c}{MMB} & \multicolumn{2}{c|}{SEED} & \multirow{2}{*}{\avgn} \\ 
        & & & &  POS & SR & OL & PR & SR & IL & \\ 
        \hline\hline

        \gray{B1} & \gray{Linear}    & \gray{144} & \gray{-} & \multicolumn{6}{c|}{\textit{\gray{Unavailable due to inflexiblity}}} & \gray{-} \\
        B2 & Resampler & 144 & 2.28 & 75.0 & 22.2 & 43.2 & 62.5 & 47.5 & 50.6 & 43.9 \\
        B3 & \cabs     & 144 & 2.23 & 135.0 & 24.4 & 54.3 & 66.7 & 49.0 & 58.8 & 53.5 \\
        
        \hline
        B4 & Linear    & 256 & 3.04 & 140.0 & 24.4 & 40.7 & 70.8 & 48.9 & 60.9 & 52.6 \\
        B5 & Resampler & 256 & 3.12 & 73.3 & 24.4 & 37.0 & 79.2 & 44.4 & 51.8 & 45.6 \\
        B6 & \cabs     & 256 & 3.07 & 136.7 & 26.7 & 55.6 & 75.0 & 52.7 & 59.3 & 56.3
        
        \end{tabular}
    }
    \end{center}
    \vspace{-0.55cm}
    \caption{\small
        \textbf{Comparison of spatial understanding capability between projectors.}
        The abbreviations for task names mean Position (POS) for MME, Spatial Relationship (SR), Object Localization (OL), and Physical Relation (PR) for MMBench, Spatial Relation (SR) and Instance Location (IL) for SEED-Bench. \avgn indicates the normalized average over six tasks. $M$ means the number of visual tokens and s/step indicates the execution time for a single step during pre-training.
    }
    \label{tab:comp_resampler}
    \vspace{-0.3cm}
\end{table}

\vspace{-0.15cm}
To showcase the value of the proposed projector, we assess and compare both performance and efficiency against existing projectors in \Cref{tab:comp_resampler} using six spatial understanding tasks from MME, MMBench, and SEED-Bench. 
First, Resampler (B2, B5) shows poor performance due to its lack of consideration for local context preservation, despite being flexible to the number of visual tokens $M$.
Second, Linear projector is limited to $M$=256 (B4) due to its inflexibility (B1), leading to intractable computational costs in high-resolution of larger $M$.
Third, for the same computational budget ($M$=256), our \cabs offer significantly improved performance compared to linear one (52.6~{\footnotesize (B4)} \vs 56.3~{\footnotesize (B6)}).
Lastly, with fewer visual tokens ($M$=144), our \cabs demonstrate improved performance (+0.9 point) and greater efficiency (3.04~{\footnotesize (B4)} \vs 2.23~{\footnotesize (B3)} s/step).
This improvement suggests our locality-enhanced projector excels at abstracting visual features where it integrates local contexts from neighboring features and provides context-enriched visual tokens, thus enabling our projectors to outperform linear counterparts even with fewer visual tokens.

\begin{table*}[!ht]
    \tablestyle{5pt}{1.05}
    \centering
    \scalebox{0.90}{
    \begin{tabular}{c|cccc|cc|cc|cc|c}
        \toprule
        \multirow{3}{*}{} & \multicolumn{6}{c|}{{Task type}} & \multicolumn{5}{c}{{MLLM benchmark}} \\ \hline
        & \multicolumn{4}{c|}{{Template-based}} & \multicolumn{2}{c|}{{GPT-assisted}} & \multicolumn{2}{c|}{{Multiple choice}} & \multicolumn{2}{c|}{{Binary yes/no}} & \multicolumn{1}{c}{{GPT eval}} \\
        & {VQA (Open)} & {VQA (MC)} & {REC} & {Cap} & {V-Inst} & {T-Inst} & \mmbdev & \seedimg & \mmep & {MME} & \llavaw \\

        \hline\hline
        D1 & \cmark & \cmark & \cmark & \cmark & \cmark & \cmark & 69.2 & 64.2 & 1568 & 1861 & 64.5 \\
        D2 & \cmark$^{*}$ & \cmark$^{*}$ & \cmark$^{*}$ & \cmark$^{*}$ & \cmark$^{*}$ & \cmark$^{*}$ & 67.4\ \footnotesize{\red{($\downarrow$1.8)}} & 63.1 & 1454\ \footnotesize{\red{($\downarrow$114)}} & 1754\ \footnotesize{\red{($\downarrow$107)}} & 62.2\ \footnotesize{\red{($\downarrow$2.3)}} \\
        \cdashline{1-12}
        D3 & & \cmark & \cmark & \cmark & \cmark & \cmark & 68.8 & 62.4\ \footnotesize{\red{($\downarrow$1.8)}} & 1310\ \footnotesize{\red{($\downarrow$258)}} & 1605\ \footnotesize{\red{($\downarrow$256)}} & 67.0 \\
        D4 & \cmark &            & \cmark & \cmark & \cmark & \cmark & 30.4\ \footnotesize{\red{($\downarrow$38.8)}} & 20.8\ \footnotesize{\red{($\downarrow$43.4)}} & 1536 & 1829 & 65.4 \\
        D5 & \cmark & \cmark &            & \cmark & \cmark & \cmark & 68.5 & 63.5 & 1524 & 1787 & 67.0 \\
        D6 & \cmark & \cmark & \cmark &            & \cmark & \cmark & 69.7 & 63.9 & 1540 & 1846 & 59.8\ \footnotesize{\red{($\downarrow$4.7)}} \\
        
        D7 & \cmark & \cmark & \cmark & \cmark &            & \cmark & 70.0 & 64.0 & 1507 & 1805 & 51.9\ \footnotesize{\red{($\downarrow$12.6)}} \\
        D8 & \cmark & \cmark & \cmark & \cmark & \cmark &            & 68.7 & 64.5 & 1559 & 1851 & 62.7\ \footnotesize{\red{($\downarrow$1.8)}} \\
        
        \cdashline{1-12}
        D9 & \cmark & \cmark & \cmark & \cmark &            &            & 70.0 & 64.5 & 1527 & 1800 & 26.1\ \footnotesize{\red{($\downarrow$38.4)}} \\
        D10 &            &            &            &            & \cmark & \cmark & 43.7\ \footnotesize{\red{($\downarrow$25.5)}} & 0.0\ \footnotesize{\red{($\downarrow$64.2)}} & 1123\ \footnotesize{\red{($\downarrow$445)}} & 1441\ \footnotesize{\red{($\downarrow$420)}} & 67.0 \\
        
        \bottomrule
    \end{tabular}
    }
    \vspace{-0.2cm}
    \caption{
    \textbf{The impact of data mixtures during instruction tuning.}
    Abbreviations for instruction data types stand for VQA (Open): open-ended visual question answering, VQA (MC): visual question answering with multiple choice, REC: referring expression comprehension, Cap: captioning, V-Inst: visual instruction, T-Inst: text-only instruction-following.
    The \cmark$^{*}$ indicates that only one dataset from each task type is used to train a model, including GQA, ScienceQA, RefCOCO, COYO100M, LLaVA150k, and ShareGPT for each task.
    }
    \label{tab:ablation:tasks}
    \vspace{-0.1cm}
\end{table*}

\vspace{-0.1cm}
\subsection{Hidden Recipe for Visual Instruction Tuning}

\definecolor{graycolor}{gray}{.873}
\newcommand\grayrow{\rowcolor[gray]{0.873}}
\newlength\savewidth\newcommand\shline{\noalign{\global\savewidth\arrayrulewidth
  \global\arrayrulewidth 1pt}\hline\noalign{\global\arrayrulewidth\savewidth}}

\begin{table*}[!t]
    \tablestyle{3pt}{1.05}
    \subfloat[
        {\textbf{Dataset balancing.} Hand-crafted balancing is the best, with per-dataset strategy serving as an effective starting point for tuning.}
        \label{table:ablation:weights}
    ]{
        \begin{minipage}{0.45\textwidth}
        \centering
        \scalebox{0.9}{
            \begin{tabular}{l|ccc|c}
            {Mixture type}    & \mmbdev & \seedimg & \mmep & \avgn  \\ \shline
            per-dataset     & 68.7             & 64.1              & 1543.2            & 70.0              \\
            per-task        & 65.7             & 62.1              & 1488.9            & 67.4              \\
            per-sample-100k  & 63.6             & 62.8              & 1494.8            & 67.1              \\
            \grayrow {per-dataset-tuned}   & \textbf{69.2}             & \textbf{64.2}              & \textbf{1568.2}            & \textbf{70.6}              \\
            \end{tabular}
        }
        \end{minipage}
    } \hspace{0.5cm}
    \subfloat[
        {
            \textbf{Instruction tuning \textit{vs.} Multi-task learning.}
            Instruction tuning (inst.) is more effective compared to multi-task learning (multi.).
        }
        \label{table:recipe:template-or-multitask}
    ]{
        \begin{minipage}{0.45\textwidth}
        \centering
        \scalebox{0.9}{
            \begin{tabular}{@{}ll|cccc|c@{}}
                Type                & Identifier    & \mmbdev  & \seedimg & \mmep  & \avgn & \llavaw \\ \shline
                \grayrow Inst.  & instruction  & \textbf{69.2} & \textbf{64.2}   & \textbf{1568.2} & \textbf{70.6} & \textbf{64.5}    \\
                Multi. & dataset name & 66.8 & \textbf{64.2}   & 1483.1 & 68.4 & 64.3    \\
                Multi. & task name    & 68.4 & 64.1   & 1507.5 & 69.3 & 64.2    \\
                \multicolumn{7}{c}{~} \\
            \end{tabular}
        }
        \end{minipage}
    }
    \\ \vspace{0.2cm}
    \subfloat[
        {
            \textbf{Template granularity and diversity.}
            The fine-grained and single template works the best for instructization.
        }
        \label{table:recipe:template-diversity-granularity}
    ]{
        \begin{minipage}{0.45\textwidth}
        \centering
        \scalebox{0.9}{
            \begin{tabular}{@{}ll|cccc|c@{}}
            Granularity & Diversity  & \mmbdev  & \seedimg & \mmep  & \avgn & \llavaw  \\
            \shline
            \grayrow fine      & single     & \textbf{69.2} & \textbf{64.2}   & 1568.2 & \textbf{70.6} & \textbf{64.5}     \\
            coarse      & single     & 68.9 & 64.0   & 1553.8 & 70.2 & 64.3     \\
            fine      & multi      & 68.1 & \textbf{64.2}   & \textbf{1581.2} & 70.5 & 61.0     \\
            fine      & multi+flip & 67.4 & 63.3   & {1575.9} & 69.8 & 62.7     \\
            \end{tabular}
        }
        \end{minipage}
    } \hspace{1cm}
    \subfloat[
        {
            \textbf{Multi-turn and de-duplication strategies.}
            Employing both strategies results in the best score.
        }
        \label{table:ablation:MT}
    ]{
        \begin{minipage}{0.45\textwidth}
        \centering
        \scalebox{0.9}{
            \begin{tabular}{cc|ccc|c}
            {MT}     & {Dedup}     & \mmbdev & \seedimg & \mmep & \avgn  \\ \shline
                            &                  & 69.1             & 63.5              & 1518.2             & 69.5  \\
            \cmark      &                  & 67.8             & 63.7              & 1546.1             & 69.6   \\
            \grayrow \cmark      &  \cmark      & \textbf{69.2}             & \textbf{64.2}              & \textbf{1568.2}             & \textbf{70.6}   \\
            \multicolumn{6}{c}{~} \\
            \end{tabular}
        }
        \end{minipage}
    }
    \vspace{-0.2cm}
    \caption{
        \small \textbf{Ablations on dataset balancing and instructization.} \avgn indicates normalized average of \mmbdev, \seedimg, and \mmep.
        Default settings are marked in \colorbox{graycolor}{gray}.
    }
    \label{table:recipes}
    \vspace{-0.3cm}
\end{table*}

\vspace{-0.15cm}
\paragraph{Dataset combination.}
\Cref{tab:ablation:tasks} shows a comprehensive ablation study to identify the individual impact of datasets on various multimodal benchmarks.
First, we investigate \textit{the impact of dataset diversity within each task} by leveraging only a single dataset for each task group (D1 \vs D2). The overall performance drop highlights the importance of the dataset diversity within each task.
Second, we explore \textit{the impact of each task} by sequentially excluding specific tasks (D1 \vs D3-8). This reveals that task diversity is crucial for learning how to handle a variety of tasks; each task improves the performance of relevant benchmarks, VQA (Open) $\rightarrow$ MME, VQA (MC) $\rightarrow$ MMB and \seedimg, and captioning and instruction-following data $\rightarrow$ \llavaw.
Third, we inspect \textit{the impact of using existing vision-language data} (D9 \vs D10). Excluding such data leads to significant decreases in MME, MMB and \seedimg benchmarks. This suggests that rich knowledge in existing vision-language datasets enhances MLLM's perception understanding or visual reasoning capabilities. In summary, these experiments emphasize the importance of diversity in both tasks and datasets within each task.

\vspace{-0.4cm}
\paragraph{Dataset balancing.}

The necessity of hand-crafted dataset balancing is addressed in previous studies \cite{flan2023,instructBLIP}. 
Based on our observations in \Cref{tab:ablation:tasks}, we tune the balance of each dataset with the two principles: limiting epochs for smaller datasets and allowing up to about a few epochs for key datasets. 
\Cref{table:ablation:weights} demonstrates the effectiveness of our manually tuned \textit{per-dataset-tuned} approach. Without hand-crafting, the \textit{per-dataset} can be a reliable alternative.
More details are provided in \cref{sec:appendix:impl-details}.

\renewcommand{\arraystretch}{0.8}
\begin{table*}[!ht]
    \tablestyle{3pt}{1.05}
    \centering
    \scalebox{0.95}{
    \begin{tabular}{l|cccc|ccccc}
        \toprule
        {Method} & {LLM} & {Projector} & {Vision Encoder} & {Res.} & \mmbdev & \mmep & {MME} & \seedimg & \llavaw \\

        \hline\hline
        \rowcolor[gray]{0.85}\multicolumn{10}{l}{\textit{\textbf{Approaches using 7B LLM}}} \\ 
        \hline        
        LLaVA (v1)~\cite{llava}         &LLaMA-7B &Linear & CLIP ViT-L/14 & 224 &38.7 &502.8 &717.5 &33.5 & - \\
        MiniGPT-4~\cite{minigpt}          &Vicuna-7B &Resampler & EVA-CLIP ViT-G & 224 &24.3 &581.7 &726.0 &47.4 & - \\
        LLaMA-AdapterV2~\cite{llamaadapterv2}    & LLaMA-7B & LLaMA-Adapter & CLIP ViT-L/14 & 224 & 41.0 &972.7 &1221.6 &32.7 & - \\
        mPLUG-Owl~\cite{mplug} &LLaMA-7B &Resampler &CLIP ViT-L/14 &224 &49.4 &967.3 &1243.4 &34.0 & - \\
        InstructBLIP~\cite{instructBLIP} &Vicuna-7B &Q-former &EVA-CLIP ViT-G &224 &36.0 &- &- &58.8 &60.9 \\
        IDEFICS &LLaMA-7B &Flamingo & OpenCLIP ViT-H/14 &224 &48.2 &- &- &44.5 &- \\
        Shikra~\cite{shikra}       &Vicuna-7B &Linear &CLIP ViT-L/14 &224 &58.8 &- &- & &- \\
        Qwen-VL~\cite{Qwen-VL} &Qwen-7B &Resampler &OpenCLIP ViT-bigG &448 &38.2 &- &- &62.3 &- \\
        Qwen-VL-Chat~\cite{Qwen-VL} &Qwen-7B &Resampler &OpenCLIP ViT-bigG &448 &60.6 &1487.5 &\underline{1848.3} &\textbf{65.4} &- \\
        LLaVA-1.5~\cite{llava-v1.5} &Vicuna-7B &Linear &CLIP ViT-L/14 &336 &64.3 &1510.7 & - & - & 63.4 \\

        \cdashline{1-10}
        \multirow{2}{*}{\model ($M$=144)} & \multirow{2}{*}{Vicuna-7B} & \cabs & \multirow{2}{*}{CLIP ViT-L/14} & \multirow{2}{*}{224} &
        \underline{70.1} &\textbf{1584.2} & \textbf{1891.3} & \underline{64.5} & \textbf{67.1} \\
        & & \dabs & & & \textbf{70.8} & \underline{1544.1} & 1835.5 & 63.8 & \underline{66.3} \\

        \midrule\hline
        \rowcolor[gray]{0.85}\multicolumn{10}{l}{\textit{\textbf{Approaches using 13B LLM}}} \\ 
        \hline
        MiniGPT-4~\cite{minigpt}    &Vicuna-13B &Resampler &EVA-CLIP ViT-G &224 & - & 866.6 & 1158.7 & - & - \\
        BLIP-2~\cite{blip2}       &Vicuna-13B &Q-former &EVA-CLIP ViT-G &224 &- &1293.8 & - & - &38.1 \\
        InstructBLIP~\cite{instructBLIP} &Vicuna-13B &Q-former &EVA-CLIP ViT-G &224 &44.0 &1212.8 &1504.6 & - &58.2 \\
        LLaVA-1.5~\cite{llava-v1.5}    &Vicuna-13B &Linear &CLIP ViT-L/14 &336 &67.7 &1531.3 &1826.7 &\underline{68.1} &70.7 \\
        \cdashline{1-10}
        \multirow{2}{*}{\model ($M$=256)} & \multirow{2}{*}{Vicuna-13B} & \cabs & \multirow{2}{*}{CLIP ViT-L/14} & \multirow{2}{*}{336} &\underline{73.2} &\underline{1629.3} &\underline{1944.0} &\textbf{68.2} & \textbf{75.7} \\
        & & \dabs & & &\textbf{73.5} &\textbf{1632.0} &\textbf{1950.0} &66.6 & \underline{72.9}\\
        \bottomrule
    \end{tabular}
    }
    \vspace{-0.2cm}
    \caption{\textbf{Comparison with other state-of-the-art MLLMs.} Res. and $M$ indicate the image resolution and the number of visual tokens, respectively. We highlight the \textbf{best results} and \underline{second-best results} in bold and underline.}
    \label{tab:comp_others}
    \vspace{-0.3cm}
\end{table*}
\renewcommand{\arraystretch}{1.0}

\vspace{-0.45cm}
\paragraph{Instruction tuning \textit{vs.} multi-task learning.}
\Cref{table:recipe:template-or-multitask} shows the advantages of instruction tuning with template-based formatting over multi-task learning using simple identifiers. This result aligns with prior studies \cite{flan,instructBLIP}, showing the efficacy of instruction tuning in our setting.

\vspace{-0.45cm}
\paragraph{Template granularity.}
\Cref{table:recipe:template-diversity-granularity} demonstrates that the fine-grained template (first row) consistently outperforms the coarse-grained template (second row) across all benchmarks. We observe that in datasets such as RefCOCO and RefCOCO+, while the input distribution $p(\mat{X}_{\texttt{img}}, \mat{X}_{\texttt{text}})$ is similar, the answer distribution $p(\mat{Y}|\mat{X}_{\texttt{img}}, \mat{X}_{\texttt{text}})$ differs. In this scenario, the coarse-grained template makes the model suffer from differentiating answers for similar inputs.

\vspace{-0.45cm}
\paragraph{Template diversity.}
To compare the effect of template diversity on model performance, we evaluate three scenarios with different diversities: using a single template (single), employing 10 templates for each dataset (multi), and inverting 3 out of 10 templates (multi+flip).
Interestingly, our experiments reveal that increasing template diversity does not guarantee a performance boost, as shown in \Cref{table:recipe:template-diversity-granularity}. This is consistent results with recent studies \cite{llava-v1.5}, showing that effective zero-shot generalization is achievable even without using multiple templates.

\vspace{-0.45cm}
\paragraph{Multi-turn template.}
\Cref{table:ablation:MT} shows the effectiveness of both multi-turn template and de-duplication strategies. The results imply removing the semantically overlapping pairs in each example is effective for mitigating shortcut training.

\vspace{-0.45cm}
\paragraph{Additional recipes.}
Apart from datasets and instructization strategies, training recipes also incorporate several subtle yet crucial design choices, including the selection of features in vision encoder, LLMs, LLM training techniques, image indicators, pre-training and instruction tuning iterations.
These recipes are detailed in \cref{sec:appendix:additional-recipes}.

\vspace{-0.45cm}
\paragraph{Final recipe.}
In summary, our \textit{final recipe} is summarized as 1) adopting flexible, locality-preserving \cabs or \dabs; 2) leveraging diverse datasets for various tasks (Table \textcolor{red}{4}); 3) applying selected ablation options in \Cref{table:recipes} and \cref{sec:appendix:additional-recipes}---the application of per-dataset balancing with hand-crafted tuning, fine-grained templates, and multi-turn interactions with deduplication.

\subsection{Putting It Altogether}
\label{sec:putting-it-altogether}

\vspace{-0.1cm}
\paragraph{Comparison with existing MLLMs.}
In \Cref{tab:comp_others}, we compare our \model, trained using the final recipe and a long training schedule, with other state-of-the-art MLLMs. Honeybee outperforms comparable 7B-scale MLLMs in all benchmarks, except for \seedimg. It is worth noting that competing methods like Qwen-VL~\cite{Qwen-VL} and LLaVA-1.5~\cite{llava-v1.5} use larger vision encoders (\eg, ViT-bigG for Qwen-VL) or larger images (448 and 336) with more visual tokens ($M$=256 and 576). In contrast, \model employs ViT-L/14 with 224 resolution and 144 visual tokens striking a balance between performance and efficiency (\Cref{fig:comp_hb_qwen_lv}). For tasks requiring detailed visual understanding, such as \seedimg (see \cref{sec:appendix:benchmark-characteristics}), using larger images or more visual tokens can be beneficial. When the number of visual tokens is increased from 144 to 256, Honeybee achieves the best score in \seedimg (65.5) among 7B-scale LLMs, as shown in \Cref{table:pushing-the-limits}.
When scaled up to 13B, \model surpasses all previous methods in every benchmark. The detailed scores are available in \cref{subsec:appendix:detailed-scores}.

\vspace{-0.4cm}
\paragraph{Pushing the limits.}
In our final 7B and 13B models, we use 144 and 256 visual tokens ($M$), respectively, balancing efficiency and performance. As indicated in \cref{fig:tradeoff} and \cref{sec:appendix:efficiency-of-mllms}, increasing $M$ consistently improves performance. 
Our experiments, aligning $M$ in \model with that of linear projector (\Cref{table:pushing-the-limits}), show performance enhancement at the cost of efficiency.
Additional comparisons with previous methods are in \cref{subsec:appendix:pushing-the-limits}.

\begin{table}[!t]
    \tablestyle{3pt}{1.05}
    \begin{center}
    \scalebox{0.9}{
        \begin{tabular}{@{}cc|cc|ccccc@{}}
        \toprule
        LLM & Res. & $M$   & s/step & \mmbdev  & \mmep  & MME  & \seedimg & \llavaw   \\
        \midrule \midrule
        \multirow{2}{*}{7B}       & \multirow{2}{*}{224}  & 144 & 2.23   & 70.1 & 1584.2 & 1891.3 & 64.5   & 67.1  \\
                 &      & 256 & 3.07   & \textbf{71.0} & \textbf{1592.7} & \textbf{1951.3} & \textbf{65.5}   & \textbf{70.6}  \\
        \midrule
        \multirow{2}{*}{13B}      & \multirow{2}{*}{336}  & 256 & 5.52   & 73.2 & 1629.3 & 1944.0 & 68.2   & 75.7  \\
                 &      & 576 & 9.80   & \textbf{73.6} & \textbf{1661.1} & \textbf{1976.5} & \textbf{68.6}   & \textbf{77.5}  \\
        \bottomrule
        \end{tabular}
    }
    \end{center}
    \vspace{-0.5cm}
    \caption{
        \small 
        \textbf{Pushing the limits} with \cabs by increasing the number of visual tokens ($M$).
        \textit{s/step} is pre-training step time.
    }
    \label{table:pushing-the-limits}
    \vspace{-0.4cm}
\end{table}

\vspace{-0.6cm}
\paragraph{Additional results.}
We additionally present (i) the detailed scores for MME, \mmbdev, \seedimg, and \llavaw  in \cref{subsec:appendix:detailed-scores}, (ii) 
ScienceQA \cite{sqa} results in \cref{subsec:appendix:sqa}, (iii) additional benchmark (MM-Vet~\cite{mmvet}, MMMU~\cite{mmmu}, POPE~\cite{pope}) results in \cref{subsec:appendix:morebenchmark}, and (iv) qualitative examples in \cref{subsec:appendix:qualitative:examples}.

\vspace{-0.15cm}
\section{Conclusion}
The advent of visual instruction tuning has brought remarkable advances in MLLMs. Despite these strides, areas such as projector design and the approach in handling multifaceted data with instructization processes remain underexplored or unclear. Inspired by this, we identify the desirable but overlooked projector property, \ie, locality preservation, and propose the locality-enhanced projector that offers a preferable performance-efficiency balance. In addition, we provide extensive experiments to identify the impact of individual design choices in handling multifaceted instruction data, unveiling hidden recipes for high-performing MLLM development. Finally, \model remarkably outperforms previous state-of-the-art methods on various benchmarks.

{
    \small
    \bibliographystyle{ieeenat_fullname}
    \bibliography{main}
}

\clearpage
\appendix

\section{Efficiency of MLLMs}
\label{sec:appendix:efficiency-of-mllms}
\begin{table}[!t]
    \tablestyle{3pt}{1.05}
    \begin{center}
    \scalebox{0.95}{
        \begin{tabular}{l|cc|ccc|c} 
        \toprule
        Projector    & $M$    & s/step & \mmbdev  & \seedimg & \mmep  & \avgn \\
        \midrule \midrule
        Linear       & 256      & 3.04      & 67.1 & 65.1   & 1556.5 & 70.0     \\ \midrule
        \multirow{4}{*}{Resampler}    & 64       & 1.69      & 65.9 & 58.9   & 1394.7 & 64.8     \\
                     & 144      & 2.28      & 66.0 & 57.0   & 1389.6 & 64.2     \\
                     & 256      & 3.12      & 67.1 & 59.9   & 1489.6 & 67.2     \\
                     & 400      & 4.27      & 67.7 & 61.5   & 1502.5 & 68.1     \\ \midrule
        \multirow{4}{*}{\cabs} & 64       & 1.65      & 69.2 & 62.9   & 1528.1 & 69.5     \\
                     & 144      & 2.23      & 69.2 & 64.2   & 1568.2 & 70.6     \\
                     & 256      & 3.07      & 70.2 & 65.3   & 1586.8 & 71.6     \\
                     & 400      & 4.15      & 70.8 & 65.5   & 1615.0 & 72.3     \\
        \bottomrule
        \end{tabular}
    }
    \end{center}
    \vspace{-0.5cm}
    \caption{
        \small 
        Detailed scores of projectors by the number of visual tokens ($M$). 
        \textit{s/step} indicates the time spent to perform one step in pre-training.
    }
    \label{table:proj-with-tokens}
\end{table}

As described in \Cref{sec:method} of the main text, the efficiency of MLLMs is predominantly affected not by the efficiency of the vision model or projector, but by the number of visual tokens (\ie, the number of output tokens of the projector). \Cref{table:proj-with-tokens} demonstrates this description, complementing \cref{fig:tradeoff}.
Notably, while the resampler has substantially larger parameters than linear (105M \vs 4M parameters), MLLM with resampler with $M=144$ is more efficient than MLLM with linear ($M=256$), as shown by lower step times (2.28 \vs 3.04).
Our \cabs, adhering to our design principles of flexibility and locality preservation, stands out as a Pareto-front model compared to both resampler and linear.

\section{Details on Projectors}
\label{sec:appendix:details-on-projectors}
In this section, we provide further ablations and descriptions for design choices of individual projectors.

\subsection{Linear Projector}
In the recent study, LLaVA (v1.5)~\cite{llava-v1.5} utilizes a 2-layer MLP instead of a single linear projection for enhancing the vision-language connector's representation power. This approach led to an investigation of how varying the number of MLP layers impacts overall performance. 
As shown in \Cref{table:proj-arch-ablation}, the 2-layer MLP-based projector marginally improves the overall performance compared to the linear projector. 
However, we observe a slight performance drop when further increasing the number of MLP layers (\ie, 6-layer MLP).
We note that our \cabs and \dabs achieve better or comparable benchmark scores while using fewer visual tokens, indicating our projectors' superiority regarding the balance of efficiency and performance.

\begin{table}[!t]
    \tablestyle{3pt}{1.05}
    \begin{center}
    \scalebox{0.95}{
        \begin{tabular}{l|ccc|c}
        \toprule
        Architectures    & \mmbdev & \seedimg & \mmep & \avgn  \\ \midrule \midrule
        Linear & 67.1 & 65.1 & 1556.5 & 70.0 \\
        2-layer MLP & 68.3 & 64.5 & 1557.2 & 70.2 \\
        6-layer MLP & 68.5 & 63.5 & 1509.2 & 69.1 \\ \midrule
        Resampler   & 66.0 & 57.0 & 1389.6 & 64.2 \\
        Resampler$_\text{w/ pos-emb}$ & 65.9 & 58.0 & 1384.7 & 64.4 \\ \midrule
        ResNet (\cabs)           & {69.2} & {64.2} & {1568.2} & {70.6} \\
        ConvNext         & 66.2 & 61.9 & 1525.4 & 68.1  \\
        StandardConv     & 67.4 & 57.1 & 1409.7 & 65.0  \\ 
        
        \midrule
        Deformable (\dabs)  & 68.6 & 63.2 & 1548.3 & 69.7 \\
        Deformable$_\text{w/o\ v-pooled Q}$ & 68.4 & 63.1 & 1521.7 & 69.2 \\
        Deformable$_\text{w/o\ M-RP}$ & 68.5 & 62.9 & 1497.0 & 68.7 \\
        \bottomrule
        \end{tabular}
    }
    \end{center}
    \vspace{-0.5cm}
    \caption{
        \small 
        Ablations for various architectural design choices in each projector.
        We use 144 visual tokens ($M$=144) for all architectures except for Linear and MLPs ($M$=256) due to their inflexibility.
    }
    \vspace{-0.2cm}
    \label{table:proj-arch-ablation}
\end{table}

\subsection{Resampler}
As described in the main text, our design focuses on two principles: 1) flexibility in visual token counts, which is the key factor to the efficiency of MLLM, and 2) preservation of local context, which is critical for spatial understanding. Our first try is augmenting visual features with positional embeddings in the resampler framework, but it does not yield notable improvements (See Resampler$_\text{w/ pos-emb}$ in \Cref{table:proj-arch-ablation}). This leads us to design two novel projectors, \cabs and \dabs.

\subsection{\cabs}
Under our design principles on flexibility and locality, we introduce convolution layers and adaptive average pooling into the projector. The overall architecture is illustrated in \cref{fig:arch}. We compare three convolution blocks: 1) ResNet bottleneck block \cite{resnet-bottleneck} with squeeze-excitation \cite{se-squeeze-excitation}, 2) ConvNext block \cite{convnext}, and 3) a standard convolution block (3$\times$3 convolution layer). \Cref{table:proj-arch-ablation} shows ResNet block outperforms ConvNext and standard convolution (StandardConv) blocks. Hence, we employ ResNet block for \cabs. While further architectural variations are explorable under the proposed design principles, we leave them for future investigation.

\newcolumntype{g}{>{\columncolor{graycolor}}l}

\begin{table*}[t]
\centering
\tablestyle{4pt}{1.05}
    \scalebox{0.95}{
        \begin{tabular}{rlll|ccccc|c}
            \toprule
        & Ablated setting      & Default value   & Changed value & \mmbdev  & \seedimg & \mmep  & MME    & \avgn & \llavaw   \\
        \midrule \midrule
        \grayrow 
        & \multicolumn{3}{c|}{\textbf{(Default)}  \model with short training schedule}                      & 69.2 & 64.2   & 1568.2 & 1860.7 & 70.6 & 64.5  \\
              \midrule
        (i)   & Image indicator      & \xmark           & \cmark             & 67.4 & 62.5   & 1543.4 & 1809.5 & 69.0 & 60.5  \\ \midrule
        (ii)  & Visual feature layer & Second-last & Last          & 69.2 & 63.7   & 1566.1 & 1839.3 & 70.4 & 62.1  \\ \midrule
        (iii) & LLM                  & Vicuna-v1.5 & LLaMA-2-chat  & 70.0 & 63.6   & 1551.7 & 1822.0 & 70.4 & 62.8  \\ \midrule
        \multirow{2}{*}{(iv)}  & \multirow{2}{*}{LLM tuning}           & \multirow{2}{*}{Full}        & LoRA ($r=64$)        & 35.0 & 48.9   & 1016.1 & 1156.1 & 44.9 & 59.2  \\
              &                      &             & LoRA ($r=256$)       & 47.3 & 49.9   & 959.1  & 1217.3 & 48.4 & 64.0  \\ \midrule \midrule
        (v)   & Pre-training steps   & 50k         & 200k          & 69.1 & 63.8   & 1586.6 & 1855.2 & 70.7 & 66.4  \\ \midrule
        \multirow{2}{*}{(vi)} & \multirow{2}{*}{Instruction tuning steps}  & \multirow{2}{*}{4k}  & 10k & 69.3 & 64.3 & 1586.8 & 1868.6 & 71.0 & 66.6  \\
             &              &     & 16k & 70.9 & 63.8 & 1550.6 & 1856.7 & 70.7 & 66.0  \\
        \bottomrule
        \end{tabular}
    }
    \vspace{-0.2cm}
    \caption{
        \textbf{Additional recipes.} 
        The default value indicates the choice used in our default ablation setting with the short training schedule. 
    }
    \label{table:additional-recipes}
    \vspace{-0.2cm}
\end{table*}

\subsection{\dabs}
We first describe how deformable attention~\cite{ddetr} works in \dabs.
The core components of deformable attention include ($i$) 2-D reference points $p$, ($ii$) 2-D sampling offsets $\Delta o$, and ($iii$) attention weights $A$. 
For individual learnable queries $\mathbf{z}$, the feature aggregation from the visual feature map $X_{feat}$ is formulated by\footnote{We recommend reading \cite{ddetr} for more details.}:
\begin{equation}
    \begin{split}
        \mathbf{z}^{l+1} =
        \sum_{k=1}^{K} A_{k}^{l}\cdot X_{feat}(p + \Delta o_{k}^{l}),
    \end{split}
    \label{eq:deform_attn}
\end{equation}
where $K$ is the number of sampling offsets per reference point, and $l$ is the index of the attention layer. All the reference points, sampling offsets, and attention weights are obtained via linear projection over the learnable queries $\mathbf{z}$; that is, they are all learnable values.
The introduction of reference points and sampling offsets for learnable queries allows locality modeling by enabling the collection of features near reference points via the sampling offsets.

On top of the deformable attention, we additionally present two techniques to improve local context modeling: 1) learnable query initialization through adaptive average pooling to the visual feature map instead of random initialization (\textit{v-pooled Q}), and 2) a manual initialization of reference points uniformly distributing on visual feature maps instead of centralized initialization (\textit{M-RP}). With these techniques, we can make reference points cover the whole region of an image, which results in offering more benefits in preserving local context with fine-grained information for a given image.
The results in \Cref{table:proj-arch-ablation} demonstrate that two techniques provide overall performance improvements of MLLMs.

\begin{table}[!t]
    \tablestyle{4pt}{1.05}
    \scalebox{0.98}{
        \begin{tabular}{l|cc}
        \toprule
        Configuration     & Pre-training & Instruction Tuning  \\
        \midrule
        Trainable modules & Abstractor   & Abstractor, LLM   \\
        Batch size        & 256          & 128               \\
        Learning rate     & 3e-4     & 2e-5                  \\
        Minimum LR        & 1e-5     & 1e-6                  \\
        LR schedule       & \multicolumn{2}{c}{Cosine decay} \\
        Warmup steps      & 2000         & 150               \\
        Training steps    & 200k         & 10k               \\
        Weight decay      & 0.01         & 1e-4              \\
        Optimizer         & \multicolumn{2}{c}{AdamW}        \\
        Optimizer HPs     & \multicolumn{2}{c}{$\beta_1=0.9,\beta_2=0.98,\epsilon=1e-6$}           \\
        Gradient clipping    & \multicolumn{2}{c}{1.0}       \\
        \bottomrule
        \end{tabular}
    }
    \vspace{-0.2cm}
    \caption{
        \small 
        \textbf{Training hyperparameters.} HP and LR indicate hyperparameter and learning rate, respectively. Note that we use LR of 1e-4 for \dabs.
    }
    \vspace{-0.7cm}
    \label{table:hyperparams}
\end{table}

\newcolumntype{?}{!{\vrule width 1pt}}

\begin{table}[!t]
    \tablestyle{3pt}{1.05}
    \begin{center}
    \scalebox{0.87}{
        \begin{tabular}{l|l|r?l|l|r}
        \toprule
        Task       & Dataset & \multicolumn{1}{l?}{Ratio} & Task & Dataset & \multicolumn{1}{l}{Ratio} \\
        \midrule\midrule
        
        \multirow[t]{4}{*}{VQA (Open)} & VQAv2 & 10.3\%  & \multirow[t]{5}{*}{REC} & RefCOCO & 10.3\% \\
        & GQA & 10.3\%  & & RefCOCO+ & 10.3\% \\
        & OCRVQA & 5.1\%  & & RefCOCOg & 10.3\% \\
        & VSR & 2.6\%  & & VG & 5.1\% \\
        \midrule
        
        \multirow[t]{2}{*}{VQA (MC)} & ScienceQA & 5.1\% & \multirow[t]{2}{*}{Instruction} & LLaVA150K & 10.3\% \\
        & A-OKVQA & 10.3\% & & ShareGPT & 2.6\% \\

        \midrule
        Captioning & COYO100M & 7.7\% & \multicolumn{1}{l}{} & \multicolumn{1}{l}{} & \\
        \bottomrule
        \end{tabular}
    }
    \end{center}
    \vspace{-0.5cm}
    \caption{
        Sampling ratio during instruction tuning. 
    }
    \label{table:dataset_weight}
\end{table}

\section{Implementation Details}
\label{sec:appendix:impl-details}
The detailed hyperparameters (HPs) are summarized in \Cref{table:hyperparams}. Additionally, we utilize total six blocks in both \cabs and \dabs (\ie, $L=3$ for \cabs and $L=6$ for \dabs in \cref{fig:arch}). We use a single node with A100 80GB $\times$ 8, employing deepspeed zero-2 \cite{zero} and flash-attention v2 \cite{flashattention2} for all experiments, except for the pre-training of long schedule where we use multi-node setups.

{
\renewcommand{\arraystretch}{1}
\begin{table*}[!t]
    \centering
    \scriptsize
    \scalebox{1.07}{
    \begin{NiceTabular}{l|l|m{12cm}}
    
    \toprule
    Task & Dataset & Template
    \\ 
    \midrule\midrule %
    
    \multirow[t]{2}{*}{Captioning} & BlipCapFilt & AI: \tmplR{caption}\\
                                \cmidrule{2-3}
                                & COYO100M & AI: \tmplR{caption} \\
    \cmidrule{1-3}

    \multirow[t]{4}{*}{VQA (Open)} & VQAv2 & Human: Answer the question using a single word or phrase. \tmpl{question} AI: \tmplR{answer}\\
                                \cmidrule{2-3}
                                
                                & GQA   & Human: Answer the question using a single word or phrase. \tmpl{question} AI: \tmplR{answer}\\
                                \cmidrule{2-3}
                                & OCRVQA & Human: Answer the question using a single word or phrase. \tmpl{question} AI: \tmplR{answer} \\
                                \cmidrule{2-3}
                                & VSR & Human: Answer the question using a single word or phrase. \tmpl{question} Please answer yes or no. AI: \tmplR{answer}\\
    \cmidrule{1-3}

    \multirow[t]{2}{*}{VQA (MC)} & ScienceQA & Human: Answer with the option's letter from the given choices directly. \tmpl{question} Context: \tmpl{context} There are several options: \tmpl{option} AI: \tmplR{answer}\\
                            \cmidrule{2-3}
                              & A-OKVQA & Human: Answer with the option's letter from the given choices directly. \tmpl{question} There are several options: \tmpl{option} AI: \tmplR{answer}\\
    \cmidrule{1-3}

    \multirow[t]{8}{*}{REC} & \multirow[t]{2}{*}{RefCOCO} & Human: Provide the bounding box coordinate of the region this sentence describes: \tmpl{phrase} AI: \tmplR{bbox}\\
                        \cmidrule{3-3}
                         &                           & Human: Provide a description for the region \tmpl{bbox}, utilizing positional words to refer to objects.  Example: `The large blue teddy bear next to the red balloon' AI: \tmplR{phrase} \\
                         \cmidrule{2-3}
                         & \multirow[t]{2}{*}{RefCOCO+} & Human: Provide the bounding box coordinate of the region this sentence describes: \tmpl{phrase} AI: \tmplR{bbox} \\
                         \cmidrule{3-3}
                         &                           & Human: Provide a description for the region \tmpl{bbox}, focusing on the appearance of objects without using positional words. Example: `The large blue teddy bear holding a red balloon.' AI: \tmplR{phrase} \\
                         \cmidrule{2-3}
                         & \multirow[t]{2}{*}{RefCOCOg} & Human: Provide the bounding box coordinate of the region this sentence describes: \tmpl{phrase} AI: \tmplR{bbox} \\
                         \cmidrule{3-3}
                         &                           & Human: Provide a description for the region \tmpl{bbox}, using detailed and descriptive expressions to refer to objects. Example: `The large blue teddy bear holding a red balloon with a joyful expression.' AI: \tmplR{phrase} \\
                         \cmidrule{2-3}
                         & \multirow[t]{2}{*}{Visual Genome} & Human: Provide the bounding box coordinate of the region this sentence describes: \tmpl{phrase} AI: \tmplR{bbox}\\
                         \cmidrule{3-3}
                         &                     & Human: Provide a short description for this region: \tmpl{bbox} AI: \tmplR{phrase} \\
    \cmidrule{1-3}
    
    \multirow[t]{2}{*}{Instruction} & LLaVA150k & Human: \tmpl{instruction} AI: \tmplR{response} \\
                                \cmidrule{2-3}
                                 & ShareGPT  & Human: \tmpl{instruction} AI: \tmplR{response} \\
    \bottomrule
    \end{NiceTabular}
    }
    \vspace{-0.15cm}
    \caption{\small
        \textbf{Templates for individual dataset.}
        We develop the templates based on LLaVA (v1.5)~\cite{llava-v1.5}.
        \{\textbf{*}\} is replaced depending on dataset examples where \textcolor{red}{red}-colored one means a target output.
        Note that \textit{bbox} is expressed as normalized coordinates $[x_{\min}$, $y_{\min}$, $x_{\max}$, $y_{\max}]$.
    }
    \label{tab:templates}
\end{table*}
}

\begin{figure*}[!t]
    \centering
    \includegraphics[width=0.85\linewidth]{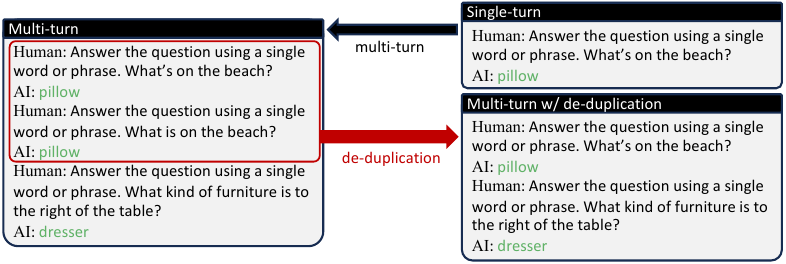}
    \vspace{-0.15cm}
    \caption{
        \textbf{The construction process of a multi-turn example with de-duplication.} This example is sampled from the GQA~\cite{gqa} dataset.
    }
    \label{fig:multiturn_dedup}
\end{figure*}

\vspace{-0.4cm}
\paragraph{Sampling ratio for datasets.}
As described in \Cref{sec:hidden_recipe}, balancing the wide range of datasets is important to train precise MLLMs. To maximize the learning of diverse knowledge from multifaceted datasets, we manually determine the sampling ratios of these datasets during training.
In pre-training, COYO100M and BlipCapFilt are used in a 1:1 ratio.
For instruction tuning, the specific sampling ratios of each dataset, determined through short schedule ablations, are detailed in \Cref{table:dataset_weight}. Notably, datasets such as VSR, ShareGPT, ScienceQA, OCRVQA, and Visual Genome (VG) have lower sampling ratios. The restricted scale of ShareGPT, VSR, and ScienceQA is due to their small dataset sizes, limited to a maximum of 3 epochs in short schedule criteria. On the other hand, the sampling ratio for OCRVQA and VG is set to 5.1\%, derived empirically from ablation experiments.
The exclusion of BlipCapFilt in instruction tuning stems from computational resource constraints, not from ablation results; we observe that including it does not notably affect the average performance.

\begin{figure*}[!ht]
    \centering
    \includegraphics[width=0.85\linewidth]{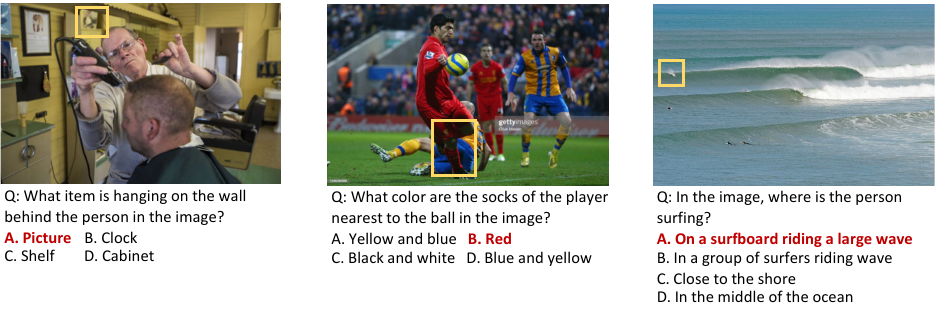}
    \vspace{-0.2cm}
    \caption{
        \textbf{Examples of SEED-Bench.} The examples require in-depth visual understanding; we highlight the regions (yellow boxes) that we need to focus on to get the correct answer (red-colored option). 
    }
    \label{fig:seed_example}
\end{figure*}
\begin{figure*}[!ht]
    \centering
    \includegraphics[width=0.9\linewidth]{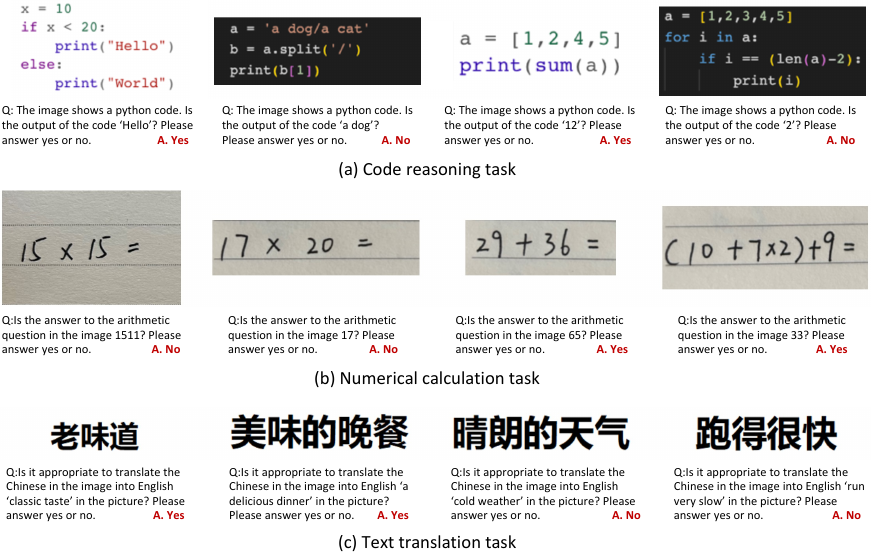}
    \vspace{-0.2cm}
    \caption{
        \textbf{Examples of MME with cognition taks.} 
    }
    \label{fig:mmecog_example}
    \vspace{-0.1cm}
\end{figure*}

\section{Additional Recipes}
\label{sec:appendix:additional-recipes}
\Cref{table:additional-recipes} presents additional ablation studies for our design choices. 
(i) There are several studies employing image indicator tokens \cite{Qwen-VL,mplug}, yet they do not demonstrate the effectiveness of the indicator tokens. Our experiments show that omitting indicator tokens improves performance.
(ii) We experiment with visual feature sources from the CLIP vision model \cite{CLIP}. The results show that utilizing features from the second-last layer rather than the last layer yields better performance \cite{blip2}.
(iii) LLaMA-2-chat and Vicuna-v1.5 show similar results, with Vicuna marginally outperforming, thus we use Vicuna. 
(iv) We applied LoRA to every query and value layer of attention following the original paper \cite{lora}, yet found full tuning of LLM to be superior. While there may be ways to better utilize LoRA, such as increasing its application scope or rank, we did not explore these further in this study.
Experiments (v) and (vi) pertain to the long training schedule employed for our final model (\Cref{tab:comp_others}).
(v) In pre-training, we freeze the LLM and train only the projector. Here, extending pre-training, a feasible option with more computational resources, is beneficial, albeit with marginal improvements.
(vi) When increasing instruction tuning steps, a broader consideration is necessary as continued LLM training can diminish its pre-trained knowledge and capabilities. Our experiments reveal that excessively long training is counterproductive, with around 10k training iterations being the most effective.

\begin{table*}[t]
\centering
\tablestyle{4pt}{1.1}
    \subfloat[
        {
            \textbf{MME scores.} Maximum scores are 200 for each subcategory, and 2000, 800, and 2800 for perception, cognition, and total, respectively.
        }
        \label{table:full-mme}
    ]{
        \centering
        \tablestyle{1.5pt}{1.1}
        \scalebox{0.8}{
            \begin{tabular}{l?cccccccccc|c?cccc|c?c}
            \toprule
                  & \multicolumn{11}{c?}{Perception}                                                                              & \multicolumn{5}{c?}{Cognition}                                                                     &   \\ \midrule
            Model & Existence & Count & Position & Color & Poster & Celebrity & Scene & Landmark & Artwork & OCR   & Sum & \makecell{Commonsense\\reasoning} & \makecell{Numerical\\calculation} & \makecell{Text\\translation} & \makecell{Code\\reasoning} & Sum & Total                 \\ \midrule
            C-7B & 185.0     & 145.0 & 161.7    & 180.0 & 166.7   & 152.4     & 157.3 & 174.5    & 129.3   & 132.5 & 1584.2     & 112.1                  & 37.5                   & 100.0             & 57.5            & 307.1     & 1891.3                \\
            D-7B & 175.0     & 153.3 & 143.3    & 175.0 & 155.4   & 148.2     & 153.3 & 163.3    & 129.8   & 147.5 & 1544.1     & 111.4                  & 47.5                   & 72.5              & 60.0            & 291.4     & 1835.5                \\
            C-13B & 185.0     & 141.7 & 173.3    & 170.0 & 178.2   & 172.4     & 160.3 & 173.5    & 142.5   & 132.5 & 1629.3     & 127.1                  & 47.5                   & 80.0              & 60.0            & 314.6     & 1944.0                \\
            D-13B & 195.0     & 175.0 & 146.7    & 168.3 & 168.0   & 164.7     & 156.5 & 174.5    & 131.0   & 152.5 & 1632.2     & 130.0                  & 62.5                   & 82.5              & 42.5            & 317.5     & 1949.7                   \\
            \bottomrule
            \end{tabular}
        }
    }
    \\ 
    \vspace{0.05cm}
    \subfloat[
        {\textbf{\seedimg accuracies.}}
        \label{table:full-seed}
    ]{
        \centering
        \scalebox{0.9}{
            \begin{tabular}{l|ccccccccc|c}
            \toprule
            Model & \makecell{Scene\\understanding} & \makecell{Instance\\identity} & \makecell{Instance\\attributes} & \makecell{Instance\\location} & \makecell{Instances\\counting} & \makecell{Spatial\\relation} & \makecell{Instance\\interaction} & \makecell{Visual\\reasoning} & \makecell{Text\\understanding} & Total  \\ \midrule
            C-7B  & 73.4                & 67.8              & 64.6                & 59.8              & 55.6               & 48.4             & 73.2                 & 74.9             & 41.2               & 64.5   \\
            D-7B  & 73.1                & 67.9              & 62.3                & 60.8              & 55.0               & 49.8             & 67.0                 & 73.1             & 27.1               & 63.5   \\
            C-13B & 75.4                & 74.0              & 68.1                & 65.5              & 59.2               & 54.2             & 71.1                 & 79.5             & 38.8               & 68.2   \\
            D-13B & 74.8                & 71.2              & 65.4                & 64.6              & 59.3               & 51.6             & 69.1                 & 78.5             & 24.7               & 66.6  \\
            \bottomrule
            \end{tabular}
        }
    }
    \\ \vspace{0.25cm}
    \subfloat[
        {\textbf{MMB accuracies.} Abbreviations stand for LR: Logic Reasoning, AR: Attribute Reasoning, RR: Relation Reasoning, FP-S: Fine-grained Perception (Single-instance), FP-C: Fine-grained Perception (Cross-instance), CP: Coarse Perception.}
        \label{table:full-mmb}
    ]{
        \begin{minipage}{0.55\textwidth}
        \centering
        \scalebox{0.9}{
            \begin{tabular}{l|cccccc|c}
            \toprule
            Model & LR   & AR   & RR   & FP-S & FP-C & CP   & Total  \\ \midrule
            C-7B  & 41.7 & 78.1 & 69.6 & 74.1 & 53.8 & 80.2 & 70.1   \\
            D-7B  & 44.2 & 75.1 & 73.0 & 73.1 & 58.6 & 81.2 & 70.8   \\
            C-13B & 45.8 & 77.6 & 77.4 & 76.8 & 57.9 & 83.6 & 73.2   \\
            D-13B & 45.0 & 75.6 & 81.7 & 76.4 & 62.1 & 82.9 & 73.5  \\
            \bottomrule
            \end{tabular}
        }
        \end{minipage}
    } \hspace{0.1cm}
    \subfloat[
        {\textbf{\llavaw scores.}}
        \label{table:full-llavaw}
    ]{
        \begin{minipage}{0.4\textwidth}
        \centering
        \scalebox{0.9}{
            \begin{tabular}{l|ccc|c}
            \toprule
            Model & Complex & Conv & Detail & All   \\ \midrule
            C-7B  & 84.6    & 50.3 & 55.1   & 67.1  \\
            D-7B  & 79.6    & 49.4 & 62.6   & 66.3  \\
            C-13B & 82.5    & 72.9 & 66.7   & 75.7  \\
            D-13B & 84.1    & 68.6 & 57.8   & 72.9  \\
            \bottomrule
            \end{tabular}
        }
        \end{minipage}
    }    
    \vspace{-0.05cm}
    \caption{
        \textbf{Detailed scores.} C- and D- in Model column indicate \cabs and \dabs, respectively. 7B and 13B indicate LLM size. For the input images, we use 224 resolution for 7B and 336 for 13B.
    }
    \label{table:full-scores}
    \vspace{-0.1cm}
\end{table*}

\section{Details on Templates}
\label{sec:appendix:template-details}
\vspace{-0.3cm}
\paragraph{Templates.}
Detailed templates for individual datasets are presented in \Cref{tab:templates}.
For captioning tasks, MLLMs are encouraged to generate directly output captions without any instructional phrase as the standard captioning task.
For VQA and REC tasks, we adopt \textit{fine-grained} templates to favorably adapt LLM's outputs for individual datasets.
For the VSR dataset, we rephrase the declarative captions into questions to suit a VQA context. For instance, a caption ``The cat is inside the refrigerator'' marked as \textit{False} is converted into ``Is the cat inside the refrigerator?'' with the answer \textit{No}.
Finally, for the instruction task, we use the original instructions and responses rather than using templates.

\paragraph{Multi-turn with de-duplication.}
For data such as VQA datasets where multiple input-target pairs exist for a single image, we make conversation-like multi-turn examples by simply concatenating the input-target pairs. Additionally, we perform a de-duplication strategy which remains only one from the duplicates (having the same target). The process is illustrated in \cref{fig:multiturn_dedup}.

\section{Benchmark Characteristics}
\label{sec:appendix:benchmark-characteristics}
Throughout this study, we observe specific characteristics in benchmarks, particularly in SEED-Bench and MME with cognition tasks (MME-cognition). SEED-Bench tends to require fine-grained visual comprehension, while MME-cognition is highly text-oriented, resulting in substantial dependency on the capabilities of LLMs. In this section, we investigate these distinctive benchmark characteristics.

\paragraph{SEED-Bench.}
We present examples of SEED-Bench, in \cref{fig:seed_example}, to show one of the major characteristics of the benchmark; we observe that the examples frequently require fine-grained visual understanding, \eg, details from small regions.
Such characteristics suggest that using large images or more visual tokens is critical in achieving higher performance in this benchmark.
Notably, in \Cref{tab:comp_others}, \model achieves competitive performance over comparative models even with smaller images or fewer visual tokens.

\paragraph{MME-cognition.}
We present examples of MME-cognition in \cref{fig:mmecog_example}.
Notably, three out of four cognition tasks are text-oriented reasoning tasks, such as code reasoning, numerical calculation, and text translation.
Consequently, the performance of these cognition tasks is predominantly influenced by which LLM is used, rather than the visual comprehension capabilities of MLLM. 
Furthermore, our analysis reveals a distinct bias in the text translation task towards Chinese-English translation. While only four examples are shown in \cref{fig:mmecog_example}, all instances of text translation tasks are observed to be Chinese-English translations. Considering such characteristics, we prioritize the MME with perception tasks (\mmep) over cognition tasks for model comparisons.

\begin{table*}[t]
    \tablestyle{4pt}{1.05}
    \scalebox{0.90}{
        \begin{tabular}{l|ccc|ccc|cc|c}
            \toprule
            \multicolumn{1}{c|}{\multirow{2}{*}{\textbf{Model}}} & \multicolumn{3}{c|}{\textbf{Subject}}        & \multicolumn{3}{c|}{\textbf{Context Modality}}       & \multicolumn{2}{c|}{\textbf{Grade}}  & \multirow{2}{*}{\textbf{Average}} \\
            \multicolumn{1}{c|}{}      & NAT & SOC & LAN & TXT & IMG & NO  & G1-6        & G7-12       &        \\ \midrule \midrule
            Human~\cite{sqa} & 90.23 & 84.97 & 87.48 & 89.60 & 87.50 & 88.10 & 91.59 & 82.42 & 88.40 \\
            GPT-3.5~\cite{sqa} & 75.44 & 70.87 & 78.09 & 74.68 & 67.43 & 79.93 & 78.23 & 69.68 & 75.17  \\
            GPT-4~\cite{llava}      & 84.06 & 73.45 & 87.36 & 81.87 & 70.75 & 90.73 & 84.69 & 79.10 & 82.69  \\
            \midrule\hline
            \rowcolor[gray]{0.85}\multicolumn{10}{l}{\textit{\textbf{Specialist Models}}} \\ 
            \hline
            LLaMA-Adapter~\cite{llamaadapter}     & 84.37 & 88.30 & 84.36 & 83.72 & 80.32 & 86.90 & 85.83 & 84.05 & 85.19  \\
            MM-CoT~\cite{mmcot}     & \textbf{95.91} & 82.00 & 90.82 & \textbf{95.26} & 88.80 & 92.89 & 92.44 & 90.31 & 91.68  \\
            LLaVA~\cite{llava}      & 90.36 & 95.95 & 88.00 & 89.49 & 88.00 & 90.66 & 90.93 & 90.90 & 90.92  \\
            LLaVA+GPT-4 $_\text{(judge)}$~\cite{llava}        & 91.56 & \textbf{96.74} & \underline{91.09} & 90.62 & 88.99 & \textbf{93.52} & 92.73 & 92.16 & 92.53  \\ \midrule\hline
            
            \rowcolor[gray]{0.85}\multicolumn{10}{l}{\textit{\textbf{Generalist Models}}} \\ 
            \hline
            Honeybee (M=256)  & 93.12 & \underline{96.63} & 90.55 & 92.52 & \underline{91.77} & 92.26 & \underline{93.72} & \underline{92.22} & \underline{93.19}  \\
            Honeybee (M=576)  & \underline{95.20} & 96.29 & \textbf{91.18} & \underline{94.48} & \textbf{93.75} & \underline{93.17} & \textbf{95.04} & \textbf{93.21} & \textbf{94.39}  \\
            \bottomrule   
        \end{tabular}
    }
    \vspace{-0.2cm}
    \caption{\small
        \textbf{Evaluation results on the Science QA test split.} Question classes: NAT = natural science, SOC = social science, LAN = language science, TXT = text context, IMG = image context, NO = no context, G1-6 = grades 1-6, G7-12 = grades 7-12. 
        Despite specialist models being tailored explicitly for the ScienceQA benchmark, \eg, further fine-tuning solely on ScienceQA, \model achieves state-of-the-art scores under a generalist approach.
        We highlight the \textbf{best results} and \underline{second-best results} in bold and underline.
    }
    \vspace{-0.15cm}
    \label{table:comp-sqa}
\end{table*}

\begin{figure}[!t]
    \begin{center}
    \scalebox{0.97}{
        \includegraphics[width=\linewidth]{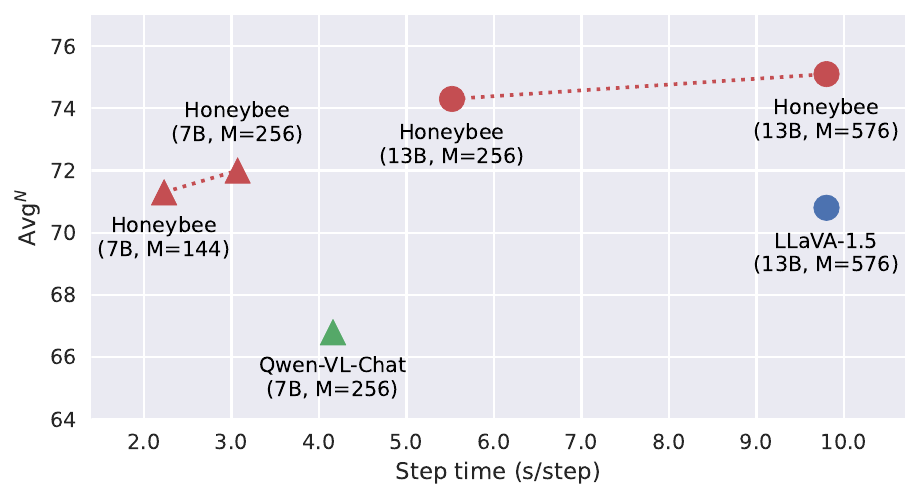}
    }
    \end{center}
    \vspace{-0.65cm}
    \caption{
        \textbf{Comparison between \model variants and current state-of-the-art methods.} \avgn denotes the normalized average score of \mmbdev, \mmep, and \seedimg.
    }
    \vspace{-0.1cm}
    \label{fig:comp_hb_qwen_lv}
\end{figure}

\section{Additional Results}
\label{sec:appendix:additional-results}
\subsection{Detailed Benchmark Scores}
\label{subsec:appendix:detailed-scores}
We report the detailed scores of our final models for all categories in MME, \mmbdev, \seedimg, and \llavaw in \Cref{table:full-scores}.

\subsection{Pushing the Limits}
\label{subsec:appendix:pushing-the-limits}

\begin{figure*}[!t]
    \begin{center}
    \scalebox{0.99}{
        \includegraphics[width=\linewidth]{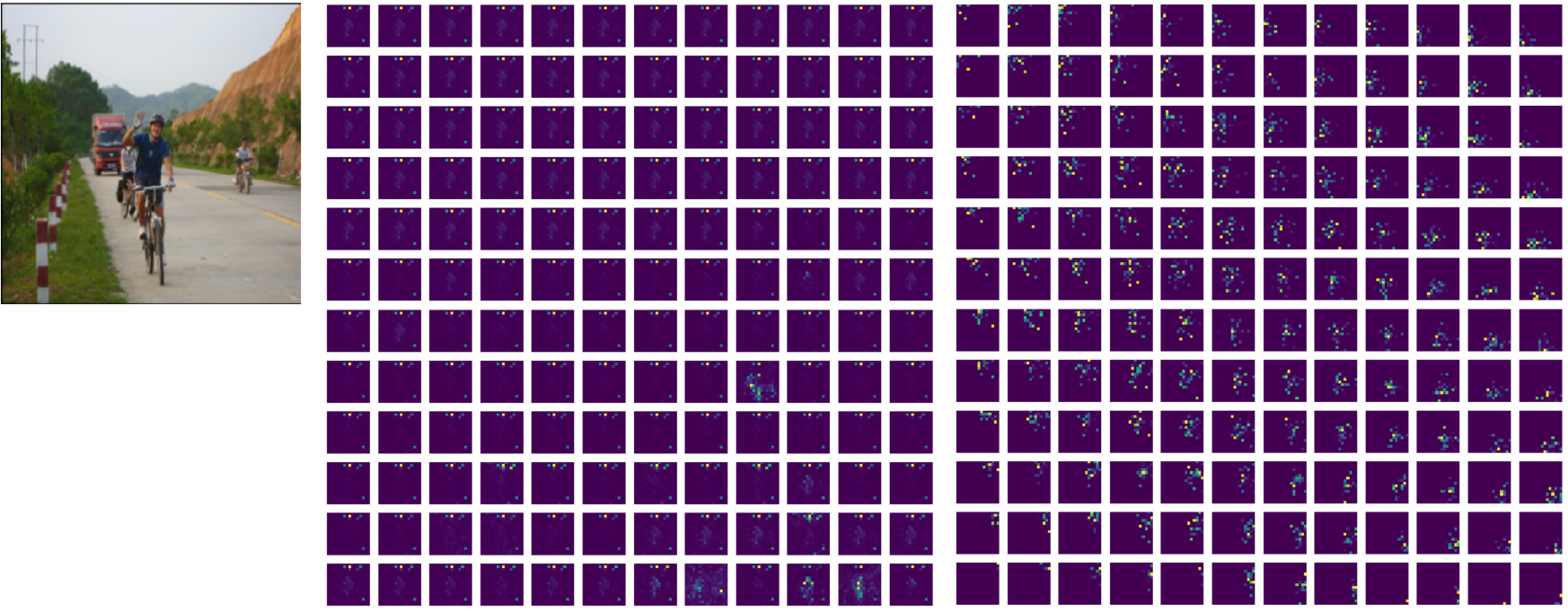}
    }
    \end{center}
    \vspace{-0.5cm}
    \caption{
         \textbf{Visualization of attention maps}. \textbf{(Left)} the input image, \textbf{(Middle)} the attention map from the resampler, and \textbf{(Right)} the attention map from \dabs.
         Our locality-aware projector (\dabs) effectively preserves local contexts, while the resampler extracts visual information mainly from a few regions and loses some details.
    }
    \label{fig:qualitative_attn}
\end{figure*}

\Cref{table:pushing-the-limits} in the main text shows the performance of \model with the increased number of visual tokens, matching them to the linear projector. Here, we further provide the comparison between the \model variants and the current state-of-the-art methods, namely Qwen-VL-Chat \cite{Qwen-VL} and LLaVA-1.5 \cite{llava-v1.5}, in \cref{fig:comp_hb_qwen_lv}.
This figure highlights the efficiency and effectiveness of the proposed \model.

\subsection{Science QA}
\label{subsec:appendix:sqa}
The Science QA dataset~\cite{sqa} is specifically designed to evaluate the broadness of domain knowledge and multi-hop reasoning skills of AI systems, which is essential for MLLMs to perform a wider range of tasks requiring more complex reasoning.
Thus, in this section, we additionally provide the evaluation results of the Science QA benchmark.  
From \Cref{table:comp-sqa}, recent MLLMs, \ie, LLaMA-adapter~\cite{llamaadapter}, MM-CoT~\cite{mmcot}, and LLaVA~\cite{llava}, show remarkable performance in this benchmark via further fine-tuning on the Science QA dataset; we refer to these fine-tuned models as \textit{Specialist Models} in \Cref{table:comp-sqa}.
Especially, in LLaVA+GPT-4 $_\text{(judge)}$, they achieved state-of-the-art scores by utilizing the GPT-4~\cite{gpt4} as a judge; whenever GPT-4 and LLaVA produce different answers, they prompt GPT-4 again, asking it to provide a final answer based on the question and two outcomes.
Remarkably, Honeybee, with \cabs and vicuna-13B, outperforms the LLaVA+GPT-4 $_\text{(judge)}$ and achieves new state-of-the-art scores in this benchmark without the assist of GPT-4 or the task-specific fine-tuning process. These results highlight the effectiveness of our contributions: 1) architectural improvement of the projector and 2) thoroughly explored training recipe.

\begin{table}[!t]
    \vspace{-0.3cm}
    \centering
    \tablestyle{3pt}{1.05}
    \scalebox{0.88}{
        \begin{tabular}{l|ccc}
        & MM-Vet        & MMMU          & POPE           \\ 
        \hline
    
        \rowcolor[gray]{0.85}\multicolumn{4}{l}{\textit{\textbf{Approaches using 7B LLM}}} \\ 
        LLaVA (v1)      & 23.8          & -                & 66.5           \\
        MiniGPT-4       & 22.1          & -                & -              \\
        LLaMA-AdapterV2 & 31.4          & 29.8             & -              \\
        mPLUG-Owl       & -             & -                & 67.4           \\
        InstructBLIP    & 26.2          & -                & -              \\
        Qwen-VL-Chat    & -             & \underline{35.9} & -              \\
        LLaVA-1.5       & 30.5          & -                & \textbf{85.9}  \\ 
        \hline
        Honeybee (C-7B-144M) & \underline{34.9} & 35.3 & 83.2 \\
        Honeybee (C-7B-256M) & \textbf{35.6} & \textbf{36.4} & \underline{84.3} \\ 
    
        \hline\hline
        \rowcolor[gray]{0.85}\multicolumn{4}{l}{\textit{\textbf{Approaches using 13B LLM}}} \\ 
        MiniGPT-4       & 24.4          & 26.8             & 74.5           \\
        BLIP-2          & 22.4          & 35.4             & 85.3           \\
        InstructBLIP    & 25.6          & 35.7             & 83.8           \\
        LLaVA-1.5       & 35.4          & \underline{36.4} & \textbf{85.9}  \\ 
        \hline
        Honeybee (C-13B-256M) & \underline{38.1} & \textbf{37.3} & 85.5 \\
        Honeybee (C-13B-576M) & \textbf{42.2} & 36.2 & \underline{85.6} 
        \end{tabular}
    }
    \vspace{-0.25cm}
    \caption{
        \small 
        \textbf{Additional benchmark comparison.} Numbers are collected from each paper and official leaderboard, selecting the best results for each method when multiple exist.
    }
    \vspace{-0.45cm}
    \label{table:morebench}
\end{table}

\subsection{Additional Benchmark Results}
\label{subsec:appendix:morebenchmark}
In addition to four benchmarks used in the main paper, we perform further evaluation using three additional benchmarks---1) MM-Vet~\cite{mmvet} and MMMU~\cite{mmmu} for visual understanding capability evaluation, and 2) POPE~\cite{pope} for evaluation of
object hallucination.
In \Cref{table:morebench}, similar to the experimental results in the main text, Honeybee shows superior comprehensive visual understanding on MM-Vet and MMMU benchmarks.
On the other hand, in the hallucination aspect, the performance of 7B-scale \model slightly falls short compared to LLaVA-1.5. However, when using larger images (336 resolution) with 13B LLM, it achieves competitive performance, suggesting the importance of higher visual understanding and reasoning to tackle hallucination issue.

\section{Qualitative Analysis}
\label{sec:appendix:qualitative}

\subsection{Attention Comparison between Resampler and D-Abstractor}
\label{subsec:appendix:qualitative:attention}
As discussed in \Cref{subsubsec:motivation}, the vanilla abstractor (\ie, resampler) tends to primarily focus on salient regions, whereas our locality-enhanced abstractor (\ie, \dabs) is designed to preserve local contexts effectively.
To further validate this, we examined attention maps from both the resampler and the D-Abstractor for their every learnable query ($M$=144).
From \cref{fig:qualitative_attn}, we observe that queries of the resampler only attend to specific salient areas, suggesting the potential loss of detailed information.
On the other hand, in the case of \dabs, each learnable query locally abstracts visual features across the whole feature map, which provides fine-grained and comprehensive information about the image.
This result indicates that our proposed projectors benefit in improving the performance of spatial understanding tasks that require to capture diverse relationships and objects in an image.

\subsection{Qualitative Examples}
\label{subsec:appendix:qualitative:examples}
\begin{figure*}[ht]
    \begin{center}
    \scalebox{0.88}{
        \includegraphics[width=\linewidth]{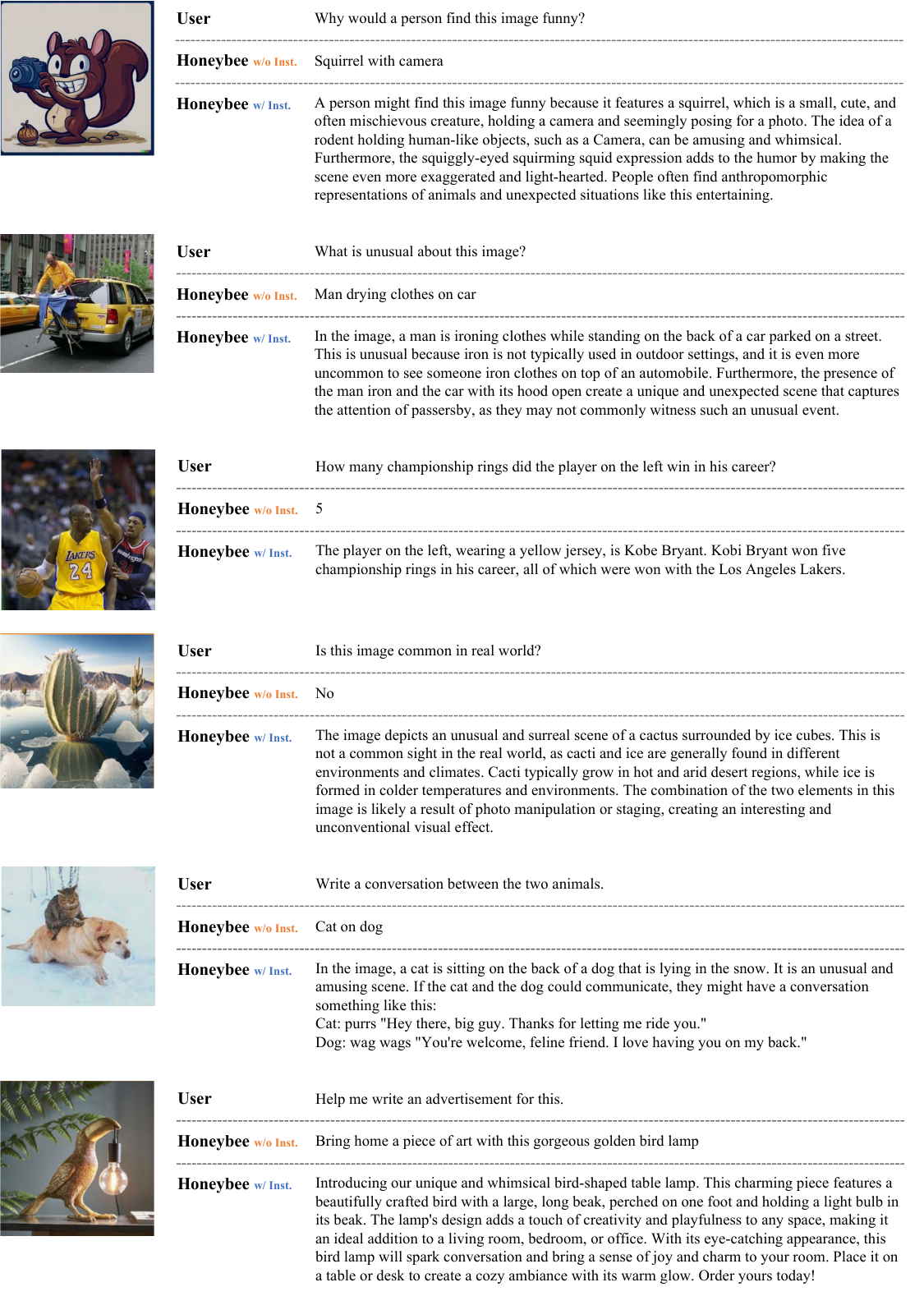}
    }
    \vspace{-0.5cm}
    \end{center}
    \caption{
         \textbf{Qualitative examples} with various user inputs. \blue{\textbf{w/ Inst.}} and \orange{\textbf{w/o Inst.}} indicate results from models trained with or without instruction-following data, \ie, LLaVA~\cite{llava} and ShareGPT~\cite{vicuna}, respectively. The example images are selected from OwlEval~\cite{mplug}.
    }
    \label{fig:qualitative_gen}
\end{figure*}

In \cref{fig:qualitative_gen}, we present generation results for various user instructions using samples of OwlEval~\cite{mplug}.
To verify the importance of instruction-following data for human preference, we report both results generated from two different models: 1) a model trained without instruction-following data such as LLaVA~\cite{llava} and ShareGPT~\cite{vicuna}, called \modelXinst (D9 in \Cref{tab:ablation:tasks}), and 2) a model trained with all available data, including the instruction-following data, called \modelOinst (D1 in \Cref{tab:ablation:tasks}).
From ~\cref{fig:qualitative_gen}, while the \modelXinst captures objects included in given images exactly, the answers tend to be short or not follow the instructions.
In contrast, \modelOinst accurately responds to user instructions and generates detailed answers preferred by humans.

\end{document}